\documentclass[letterpaper]{article}
\usepackage[preprint]{aaai2027}
\usepackage[hyphens]{url}
\usepackage{graphicx}
\usepackage{booktabs}
\usepackage{amsmath,amssymb,multirow}
\usepackage{natbib}
\usepackage{caption}
\title{When Watermarks Turn Adversarial: Exposing Vulnerabilities in Document Understanding}
\author{
Qingzhe Han, Fan Dang, Shengqi Chen, Xinyi Sun,\\
Haoyu Xiong, Lingkun Li, Di Zhang
}
\affiliations{Beijing Jiaotong University}

\newcommand{\method}{WASP}
\begin{document}
\maketitle
\begin{abstract}
Visible watermarks serve legitimate purposes in digital documents and often remain present during vision--language model (VLM) processing. However, these ordinary visual elements may also become adversarial surfaces that compromise document understanding. We investigate this vulnerability and propose Watermark-based Adversarial Surface Perturbation (WASP), a white-box attack that disrupts visual representations by optimizing pixels within fixed watermark strokes, without requiring downstream questions, answers, or task-specific losses. We further introduce Balanced Aggregation Loss (BAL), which emphasizes visual token groups retaining high similarity to clean representations.

Experiments show that adversarial watermarks cause substantially greater degradation than ordinary watermarks. On DocVQA, WASP achieves attack success rates of 96.81\% and 71.93\% against Qwen2.5-VL and InternVL2.5, respectively, with BAL providing gains of 9.21 and 2.24 percentage points over a standard token hinge objective. On 1,651 OmniDocBench pages, WASP increases MinerU2.5's text edit distance by 0.7748 relative to ordinary watermarks. Beyond direct white-box attacks, WASP can exhibit strong transfer within model families and between readers with the same vision-tower architecture. Further analyses reveal that attack effectiveness extends beyond direct text occlusion and depends on watermark geometry, model architecture, and preprocessing. In structured document parsing, disrupted layout and routing decisions can propagate errors to downstream extraction. These findings expose a previously underexplored security risk in document understanding systems.
\end{abstract}

\section{Introduction}
\label{sec:intro}

% Final user-supplied vector overview; whitespace cropped in the PDF asset.
% Preserve the artwork aspect ratio when fitting to the full text width.
\begin{figure*}[t]
\centering
\includegraphics[width=\textwidth,keepaspectratio]{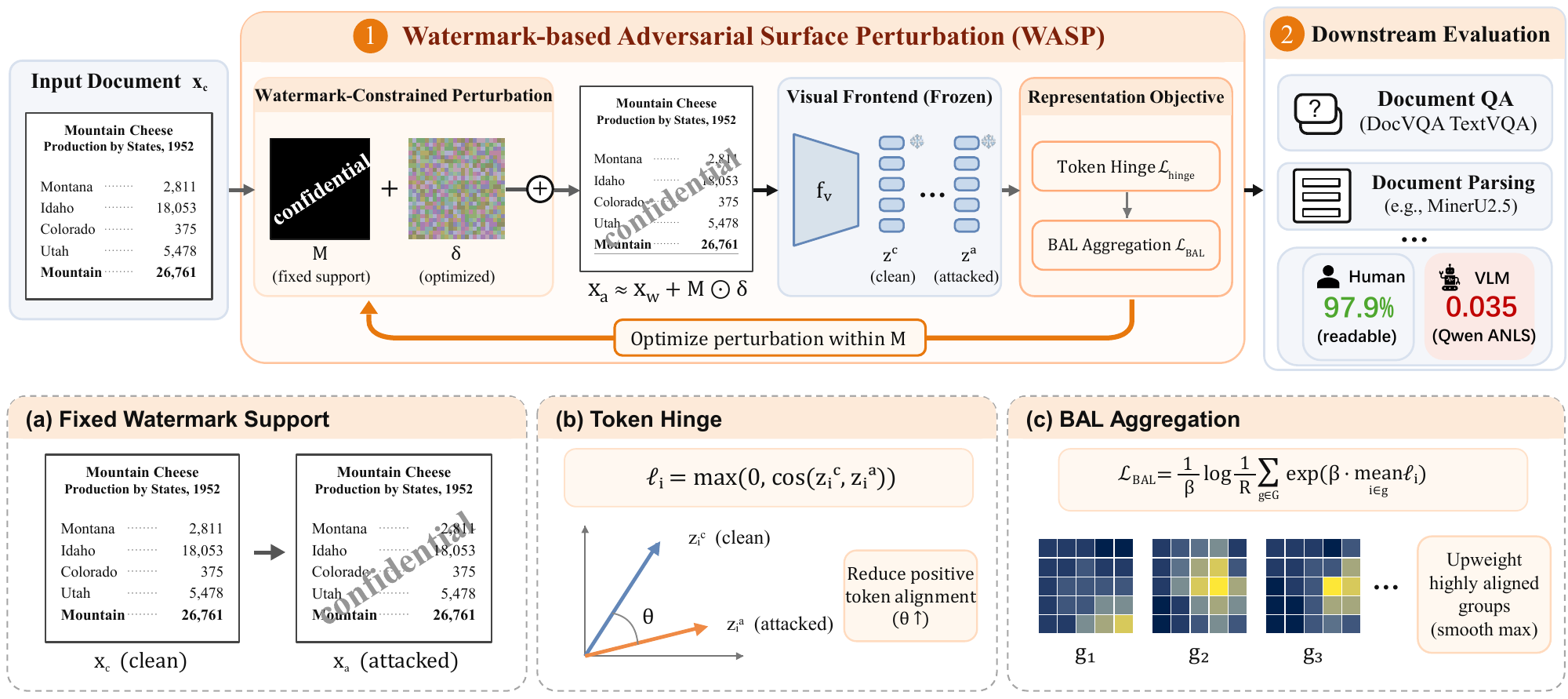}
\caption{Overview of \method. A repeated text watermark defines a fixed editable support. Optimization modifies its stroke pixels through a frozen visual frontend and a representation objective. The exported page is evaluated on downstream document tasks.}
\label{fig:overview}
\end{figure*}

Vision--language models (VLMs) have made document question answering and structured parsing increasingly practical. Yet real documents contain more than the text, tables, and figures that these systems are expected to understand. Visible watermarks, for example, communicate ownership, confidentiality, or document status, and may be retained when documents are shared or automatically processed. Unlike incidental visual noise, these marks have a legitimate role, so their presence cannot always be treated as an artifact to remove. This creates a security tension: \emph{can a visual element that document understanding systems are expected to tolerate become an adversarial surface that compromises their interpretation of the document?}

Adversarial perturbations and patches can disrupt visual recognition~\cite{goodfellow2015explaining,brown2017patch}; more recent attacks show that perturbing visual representations can compromise downstream VLM tasks without task-specific labels~\cite{zhou2023advclip,zhang2025anyattack,mei2025veattack}. In document understanding, ordinary visible watermarks have been shown to affect VLM performance~\cite{xu2025watermark}, while adversarial watermark attacks have been studied for OCR systems~\cite{chen2020fawa}. What remains insufficiently understood is how a watermark can act as an \emph{active} adversarial carrier in modern document understanding, rather than merely obstructing text. Repeated text watermarks are particularly interesting: their glyphs occupy a limited, spatially disconnected set of pixels, yet extend across much of a page. As a result, the number of editable pixels or the amount of text directly covered may not capture their influence on a VLM. That influence may also differ across visual frontends and document processing pipelines.

We investigate three connected aspects of this risk. First, can adversarially optimized watermarks induce failures substantially beyond those caused by ordinary watermarks, without targeting particular questions or answers? Second, how do watermark geometry, placement, and model architecture shape their effects, including when the watermark largely avoids document content? Third, can the resulting errors propagate through intermediate stages of document parsing, rather than appearing only as incorrect final answers? These questions frame visible watermarks as a document-specific adversarial surface whose effectiveness and failure modes warrant systematic study.

To study this setting, we introduce \textbf{Watermark-based Adversarial Surface Perturbation (WASP)}. The attack renders a repeated text watermark, fixes its glyph geometry and placement, and restricts subsequent pixel modifications to the watermark strokes. This fixed support separates the effect of adversarial optimization from that of the ordinary watermark itself. Building on VEAttack\textquotesingle s representation-level attack principle~\cite{mei2025veattack}, WASP confines visual-token disruption to fixed watermark strokes, using white-box access to the visual frontend without downstream questions, answers, or task losses. We further introduce \textbf{Balanced Aggregation Loss (BAL)}, a grouped hinge objective that places greater weight on token groups remaining highly aligned with the clean document. Unlike a global token average, this aggregation discourages optimization from neglecting less-disrupted groups.

We evaluate WASP on document VLMs including Qwen2.5-VL~\cite{bai2025qwen} and InternVL2.5~\cite{chen2025internvl}, using DocVQA~\cite{mathew2021docvqa} and TextVQA~\cite{singh2019towards}. On the DocVQA validation split, WASP achieves attack success rates of \textbf{96.81\%} and \textbf{71.93\%} against Qwen2.5-VL and InternVL2.5, respectively. BAL improves these rates by 9.21 and 2.24 percentage points over a token hinge objective. Controls with ordinary watermarks and amplitude-matched random perturbations, together with carrier and placement studies, show that the observed degradation cannot be attributed solely to passive occlusion or perturbation magnitude. Notably, background-prioritized watermarks remain highly effective against Qwen2.5-VL, whereas InternVL2.5 is more sensitive to placement near text, revealing marked model dependence.

We also evaluate structured parsing with MinerU2.5~\cite{wang2025mineru} on 1,651 OmniDocBench pages~\cite{ouyang2025omnidocbench}. WASP increases text edit distance by \textbf{0.7748} relative to ordinary watermarks. A routing intervention provides a more specific insight: restoring clean layout and routing decisions largely recovers text extraction from attacked content, whereas retaining attacked routing decisions causes severe errors even with clean recognition content. This points to upstream layout and routing as an important failure pathway in the evaluated parser, beyond incorrect recognition of isolated characters. Beyond the source model, the optimized watermarks can exhibit strong transfer within model families and between readers with the same vision-tower architecture, extending their impact to readers that were not used during optimization. This finding broadens the security implications of adversarial watermarks: an overlay optimized against one reader can also compromise architecturally related document understanding systems.

Our contributions are threefold:
\begin{itemize}
    % \item We identify and systematically study \textbf{visible watermarks as adversarial surfaces} in document understanding, distinguishing active attacks from ordinary watermark interference under a fixed watermark support.

    % \item We introduce \textbf{WASP and BAL}, a question-independent, watermark-constrained representation attack and a group-aware optimization objective, and quantify their effectiveness across document VLMs.

    % \item We analyze \textbf{geometry, model dependence, and failure propagation} through controlled carrier and placement experiments, transfer evaluations, and document-parsing interventions, demonstrating strong transfer in architecturally related model pairs and vulnerabilities in upstream layout and routing decisions.
    
    \item We establish visible text watermarks as an
    adversarial surface for document understanding.
    Through fixed-support controls, we demonstrate
    that adversarial optimization causes substantially
    greater degradation than ordinary watermark
    interference.

    \item We introduce WASP, a question-independent
    attack that constrains representation perturbations
    to fixed watermark strokes, and BAL, a group-aware
    objective that improves attack effectiveness by
    emphasizing regions with high residual feature
    similarity.

    \item We systematically investigate watermark
    geometry, placement, model dependence, and
    transferability. In structured parsing, controlled
    routing interventions reveal how upstream
    layout failures propagate to downstream extraction.
\end{itemize}

\section{Related Work}
\paragraph{Watermarks and document robustness.}
Watermarking includes visible marks for ownership and attribution~\cite{dekel2017watermarks} and invisible signals for data hiding~\cite{zhu2018hidden}. In document workflows, visible marks also communicate confidentiality and revision status. FAWA demonstrates adversarial watermark attacks on OCR systems~\cite{chen2020fawa}. Xu et al.~\cite{xu2025watermark} characterize ordinary watermark interference; our controls hold watermark support fixed and show substantially greater degradation from adversarial optimization (Table~\ref{tab:qa}). Counterfeit Answers~\cite{pintore2025counterfeit} optimizes answer-level losses for known questions, whereas WASP optimizes visual representations within fixed watermark strokes without downstream questions, answers, or task losses.

\paragraph{Constrained adversarial attacks.}
Early work established the sensitivity of neural networks to small input perturbations~\cite{szegedy2014intriguing,goodfellow2015explaining}. Iterative and optimization-based attacks provide stronger robustness evaluations~\cite{madry2018towards,carlini2017towards}, while momentum improves adversarial transferability~\cite{dong2018boosting}. Adversarial patches~\cite{brown2017patch} and physical-world attacks~\cite{eykholt2018robust} show that localized modifications can survive practical image transformations. Our constraint comes from the geometry of a conventional watermark: character holes and spaces between letters remain outside the editable domain. This support is fixed before optimization, so its location and area can be held constant when comparing objectives.

\paragraph{Attacks on vision--language representations.}
Set-level guidance~\cite{lu2023set} and VLATTACK~\cite{yin2023vlattack} exploit pretrained vision--language models to attack downstream tasks. AdvCLIP~\cite{zhou2023advclip} and AnyAttack~\cite{zhang2025anyattack} study downstream-agnostic or self-supervised attacks, and VEAttack~\cite{mei2025veattack} directly reduces the similarity between clean and attacked visual tokens. Our study builds on this principle through a watermark-constrained attack formulation and group-aware aggregation for uneven regional disruption, with a central focus on systematically characterizing vulnerabilities in document understanding and parsing.

Visual attacks also expose broader vulnerabilities in generative VLMs~\cite{zhao2023evaluating}. Image Hijacks~\cite{bailey2024hijacks} studies adversarial control of generation, and cross-prompt attacks examine reuse across instructions~\cite{luo2024lies}. Visual jailbreaks~\cite{qi2024visual}, compositional attacks~\cite{shayegani2024pieces}, and typographic prompts such as FigStep~\cite{gong2025figstep} target instruction following or safety alignment. Our evaluation concerns errors in document understanding: attack generation uses visual features without optimizing an answer, an instruction, or a jailbreak response.

\paragraph{\mbox{Document understanding and evaluation.}}
DocVQA~\cite{mathew2021docvqa} evaluates question answering on document images. Qwen2.5-VL~\cite{bai2025qwen} and InternVL2.5~\cite{chen2025internvl} provide complementary visual frontends for this task. OmniDocBench~\cite{ouyang2025omnidocbench} evaluates structured parsing, including text, formulas, tables, and reading order. MinerU2.5~\cite{wang2025mineru} separates coarse layout analysis from content recognition, allowing us to inspect how perturbations propagate through routing decisions. We report task-specific metrics for QA and parsing, alongside controls on watermark support and optimization.

% BEGIN retained DocOwl text (not rendered)
% DocVQA~\cite{mathew2021docvqa} evaluates question answering on document images. Qwen2.5-VL~\cite{bai2025qwen}, InternVL2.5~\cite{chen2025internvl}, and DocOwl2~\cite{hu2024docowl2} provide complementary visual frontends for this task. OmniDocBench~\cite{ouyang2025omnidocbench} evaluates structured parsing, including text, formulas, tables, and reading order. MinerU2.5~\cite{wang2025mineru} separates coarse layout analysis from content recognition, allowing us to inspect how perturbations propagate through routing decisions. We report task-specific metrics for QA and parsing, alongside controls on watermark support and optimization.
% END retained DocOwl text

\section{Watermark-based Adversarial Surface Perturbation}
\label{sec:method}
WASP confines representation-level attacks to fixed watermark strokes (Figure~\ref{fig:overview}). It builds on VEAttack's visual-token disruption principle~\cite{mei2025veattack}, with BAL addressing uneven disruption across visual regions.

\subsection{Threat model and watermark constraint}
The attacker modifies pixels within a fixed text watermark, with white-box access to the frozen visual frontend, connector, and native preprocessing. Optimization and checkpoint selection use no downstream questions, answers, losses, or QA scores.

Let $x_c\in[0,1]^{H\times W\times3}$ be the clean page, $x_w$ its ordinary-watermarked version, and $M\in\{0,1\}^{H\times W}$ the fixed stroke mask (Figure~\ref{fig:overview}a). To constrain perturbations to this carrier, WASP constructs
\begin{equation}
 x_a(\omega)=\operatorname{clip}_{[0,1]}\left(x_w+M\odot\delta(\omega)\right),
 \quad \delta(\omega)=\epsilon\tanh\omega.
 \label{eq:param}
\end{equation}
Here $H,W$ are page height and width, $\odot$ denotes elementwise multiplication, and $\omega$ is a single-channel field broadcast across RGB, initialized at zero so that $x_a(0)=x_w$. The bound $\epsilon$ controls perturbation amplitude, $M$ preserves pixels outside the strokes, and clipping enforces the valid image range.

\subsection{Visual-token disruption}
To disrupt features without task-specific supervision, we reduce alignment between corresponding unit-normalized clean and attacked tokens $z_i^c$ and $z_i^a$ from the frozen frontend $f_v$ and its connector. The token hinge in Figure~\ref{fig:overview}b gives
\begin{equation}
 \ell_i=\max\left(0,\cos(z_i^c,z_i^a)\right),
 \quad \mathcal L_{\mathrm{hinge}}=\frac{1}{N}\sum_{i=1}^{N}\ell_i.
 \label{eq:hinge}
\end{equation}
The hinge reduces positive alignment until cosine similarity reaches zero. Clean features are fixed references; gradients update the watermark pixels through attacked features. The Hinge baseline averages all $N$ valid image tokens, including thumbnails.

\subsection{Balanced Aggregation Loss}
A global average can conceal regions that remain highly aligned while most tokens are already disrupted. BAL emphasizes these regions by aggregating group-level residual alignment with a smooth maximum (Figure~\ref{fig:overview}c).

For a fixed partition $\mathcal G=\{\mathcal G_1,\ldots,\mathcal G_R\}$ of valid tokens into $R$ nonempty groups, let $s_r$ be the mean hinge penalty in group $r$ and define
\begin{align}
 s_r&=\frac{1}{|\mathcal G_r|}\sum_{i\in\mathcal G_r}\ell_i,\nonumber\\
 \mathcal L_{\mathrm{BAL}}&=\frac{1}{\beta}\log\left(\frac{1}{R}\sum_{r=1}^{R}\exp(\beta s_r)\right).
 \label{eq:bal}
\end{align}
We use log-sum-exp as a smooth approximation to the maximum, prioritizing the least-disrupted groups while distributing optimization weights across groups.
The inverse temperature $\beta>0$ controls emphasis on high-scoring groups: their gradient weights are proportional to $\exp(\beta s_r)$.

Groups are constructed once from the clean page and native frontend geometry. Qwen uses mutually exclusive layout regions and background, with layout predictions from a frozen MinerU2.5 detector. InternVL uses its local tiles and thumbnail, and MinerU uses a spatial grid. Grouping details appear in the supplement.

\subsection{Optimization and downstream evaluation}
To combine representation disruption with control of pixel changes and spatial variation, we optimize
\begin{align}
 \mathcal L&=\lambda_{\mathrm{rep}}F+\lambda_{\mathrm{vis}}V,
 \qquad F=k_B\mathcal L_{\mathrm{BAL}},\label{eq:loss}\\
 V&=\frac{\sum_{p,c}M_p(x_a(\omega)_{pc}-x_{w,pc})^2}{\max(1,\sum_pM_p)}
       +\operatorname{TV}(x_a(\omega)).\nonumber
\end{align}
Here $F$ is the representation objective, $V$ is the visual regularizer, and the $\lambda$ coefficients weight their contributions; $p,c$ index pixels and RGB channels. The squared-error term controls changes within the strokes. Total variation (TV) sums the mean absolute differences between adjacent pixels in each spatial direction over the full image, encouraging smoothness. The fixed, model-specific scale $k_B$ matches BAL's initial within-mask gradient magnitude to that of the global hinge objective. The Hinge ablation uses $F=\mathcal L_{\mathrm{hinge}}$ under the same optimization budget.

\paragraph{Implementation settings.}
The default watermark repeats lowercase \texttt{confidential} at 24-pixel type, $45^\circ$ rotation, and opacity 0.30, covering approximately 8\% of the page. We set $\epsilon=0.5$, $\beta=10$, $\lambda_{\mathrm{rep}}=0.5$, and $\lambda_{\mathrm{vis}}=0.02$, and use Adam with learning rate 0.01 for 250 updates. Native preprocessing operates on quantized pixels, with a straight-through gradient approximation for nondifferentiable operations. Carrier construction, gradient calibration, and numerical details are provided in the supplement.

We export the page with the lowest complete loss $\mathcal L$ over updates 1--250. Each ablation arm selects using its own objective. Auxiliary studies retain their stated 300-update protocols. The exported page is reused across questions or passed to a parser. For MinerU2.5, optimization targets layout-stage image tokens, and the resulting page passes through the full parser, including layout analysis, routing, crop recognition, and assembly.

\section{Experiments}
\subsection{Experimental setup}
\paragraph{Data and models.}
Both Qwen2.5-VL-7B and InternVL2.5-8B use the same datasets: the complete DocVQA validation split (1,285 pages, 5,349 questions) and a fixed TextVQA validation sample (1,000 images, 1,592 questions). Trajectory, carrier, and placement analyses use 200 uniformly sampled DocVQA pages (894 questions; seed 20261028). Each attacked image is reused for all associated questions. Models are frozen and retain native preprocessing; decoding is greedy with a 256-token limit and a fixed concise-answer prompt.

Structured parsing uses MinerU2.5 on all 1,651 OmniDocBench v1.6 pages. We attack layout-stage visual representations and evaluate the complete native parser.

\paragraph{Metrics and statistical units.}
DocVQA uses ANLS and exact match (EM); TextVQA uses official VQA accuracy. Strict semantic ASR is measured on questions answered correctly by each model on clean images. A fixed judge counts wrong-key, off-topic, and no-answer outcomes as attack successes. Clean ASR is undefined. The supplement gives scoring details and endpoint counts.

Confidence intervals use 10,000 paired page-bootstrap resamples (seed 20261028), moving all questions together and recomputing clean-correct denominators; 95\% intervals are unadjusted for multiple comparisons. Parsing uses official text, formula, and reading-order edit distance (ED; higher is worse), table TEDS and formula CDM (lower is worse), with official category-missing rules. Secondary intervals appear in the supplement.

\paragraph{Comparison settings.}
We compare Clean, ordinary watermarks (Plain), random perturbations (Random), and WASP. Plain, Random, and WASP share stroke masks. Random averages three seeds targeting WASP's within-mask RMS after clipping and quantization; matching exceptions appear in the supplement. Main DocVQA, TextVQA, and Hinge/BAL comparisons use 250 updates with loss-selected endpoints. Auxiliary studies retain their stated budgets and endpoints. The supplement reports distortion audits and generation costs for newly computed 250-update attacks, excluding reused images. Bold values in attack-comparison tables indicate the strongest attack within each matched comparison, without implying statistical significance.

\subsection{Adversarial Effects beyond Ordinary Watermarks}
\begin{table}[t]
\centering\small
\caption{DocVQA, 250 updates, loss-selected WASP with BAL aggregation. Both models use the same validation split of 1,285 pages and 5,349 questions. Random averages three same-mask, RMS-targeted noise seeds. EM and ASR are percentages.}
\label{tab:qa}
\setlength{\tabcolsep}{4pt}
\begin{tabular}{llrrr}
\toprule
Model & Input & ANLS $\downarrow$ & EM $\downarrow$ & ASR $\uparrow$ \\
\midrule
Qwen2.5-VL & Clean & 0.9310 & 86.51 & --- \\
 & Plain & 0.9101 & 82.70 & 4.01 \\
 & Random & 0.9034 & 81.81 & 5.21 \\
 & WASP & \textbf{0.0345} & \textbf{2.03} & \textbf{96.81} \\
\midrule
InternVL2.5 & Clean & 0.9169 & 85.38 & --- \\
 & Plain & 0.8639 & 77.34 & 8.90 \\
 & Random & 0.8730 & 77.98 & 8.29 \\
 & WASP & \textbf{0.3084} & \textbf{18.32} & \textbf{71.93} \\
\bottomrule
\end{tabular}
\end{table}

% BEGIN retained DocOwl table (not rendered)
% \begin{table}[t]
% \centering\small
% \caption{DocVQA, 250 updates, loss-selected AWM (hinge+BAL). All conditions share the evaluated pages within each model: Qwen 1,282 pages / 5,330 questions; InternVL and DocOwl2 1,285 / 5,349 each. Random averages three same-mask, RMS-targeted noise seeds. ASR denominators are 5,081/4,919/4,125; EM and ASR are percentages.}
% \label{tab:qa}
% \setlength{\tabcolsep}{4pt}
% \begin{tabular}{llrrr}
% \toprule
% Model & Input & ANLS $\downarrow$ & EM $\downarrow$ & ASR $\uparrow$ \\
% \midrule
% Qwen2.5-VL & Clean & 0.9310 & 86.51 & --- \\
%  & Plain & 0.9101 & 82.70 & 4.01 \\
%  & Random & 0.9034 & 81.81 & 5.21 \\
%  & AWM & 0.0345 & 2.03 & 96.81 \\
% \midrule
% InternVL2.5 & Clean & 0.9169 & 85.38 & --- \\
%  & Plain & 0.8639 & 77.34 & 8.90 \\
%  & Random & 0.8730 & 77.98 & 8.29 \\
%  & AWM & 0.3084 & 18.32 & 71.93 \\
% \midrule
% DocOwl2 & Clean & 0.7788 & 68.78 & --- \\
%  & Plain & 0.6667 & 54.81 & 21.31 \\
%  & Random & 0.6746 & 55.31 & 20.07 \\
%  & AWM & 0.1503 & 8.26 & 85.33 \\
% \bottomrule
% \end{tabular}
% \end{table}
% END retained DocOwl table

\paragraph{Text-centric QA: DocVQA.}
Table~\ref{tab:qa} shows substantial degradation beyond ordinary watermarks: ANLS falls from Plain's 0.9101/0.8639 to 0.0345/0.3084 for Qwen/InternVL. WASP reaches 96.81\%/71.93\% ASR, exceeding Plain by 92.80/63.02 percentage points and same-mask Random (5.21\%/8.29\%) by large margins. With fixed stroke support and comparable within-mask magnitude, these gaps support the role of adversarial optimization beyond passive coverage and the tested random directions.

\paragraph{Text-centric QA: TextVQA.}
\begin{table}[t]
\centering\small
\caption{TextVQA, 250 updates. Both models use the same validation sample of 1,000 images and 1,592 questions. ASR is measured on each model's clean-correct set. Random averages three seeds.}
\label{tab:textvqa_main}
\begin{tabular}{llrr}
\toprule
Model & Input & VQA acc. $\downarrow$ & ASR (\%) $\uparrow$ \\
\midrule
Qwen2.5-VL & Clean & 0.7948 & ---\\
 & Plain & 0.7633 & 5.68\\
 & Random & 0.7541 & 5.98\\
 & WASP & \textbf{0.3065} & \textbf{58.31}\\
\midrule
InternVL2.5 & Clean & 0.7676 & ---\\
 & Plain & 0.7091 & 7.40\\
 & Random & 0.6948 & 8.37\\
 & WASP & \textbf{0.2162} & \textbf{55.87}\\
\bottomrule
\end{tabular}
\end{table}

On scene-text images, WASP reduces official VQA accuracy from 0.7633 to 0.3065 for Qwen and from 0.7091 to 0.2162 for InternVL relative to Plain (Table~\ref{tab:textvqa_main}). ASR reaches 58.31\% and 55.87\%, respectively, extending the additional damage beyond ordinary watermarks to TextVQA. Same-mask Random yields only 5.98\% and 8.37\% ASR. Paired uncertainty estimates and coverage details appear in the supplement.

\begin{table}[t]
\centering\scriptsize
\caption{Human readability: 15 volunteers, 750 image ratings per condition. Doc./Mark are pooled affirmative percentages, with affirmative/total counts in parentheses; $n$ denotes volunteers. Clean Mark is not applicable.}
\label{tab:human}
\setlength{\tabcolsep}{2pt}
\begin{tabular}{lrrr}
\toprule
Input & Doc. (\%) & Mark (\%) & Ratings \\
\midrule
Clean & 99.2 (744/750) & N/A & 750 ($n=15$) \\
Plain watermark & 98.4 (738/750) & 95.6 (717/750) & 750 ($n=15$) \\
WASP--Qwen & 97.9 (734/750) & 94.8 (711/750) & 750 ($n=15$) \\
WASP--InternVL & 97.6 (732/750) & 94.3 (707/750) & 750 ($n=15$) \\
\bottomrule
\end{tabular}
\end{table}

% BEGIN retained DocOwl table (not rendered)
% \begin{table}[t]
% \centering\small
% \caption{Planned formal evaluation of source-content reading and watermark recognition (results pending). Cells are intentionally empty; qualitative manual inspection is described in the text. Read: human QA accuracy; Time: median human QA response time, excluding watermark transcription; Mark: watermark-word transcription accuracy; $n$: completed page--question ratings. Clean has no watermark, so Mark is not applicable.}
% \label{tab:human}
% \setlength{\tabcolsep}{4pt}
% \begin{tabular}{lrrrr}
% \toprule
% Input & Read (\%) & Time (s) & Mark (\%) & $n$ \\
% \midrule
% Clean & & & & \\[2pt]
% Plain watermark & & & & \\[2pt]
% AWM--Qwen & & & & \\[2pt]
% AWM--InternVL & & & & \\[2pt]
% AWM--DocOwl2 & & & & \\[2pt]
% \bottomrule
% \end{tabular}
% \end{table}
% END retained DocOwl table

% Illustrative numbers ONLY, NOT measured results; kept out of the PDF.
% Assumption: 200 unique images, one question and one rating per image per condition.
% Clean             & 97.0 & 8.1 & ---  & 200 \\
% Plain watermark   & 96.5 & 8.4 & 98.5 & 200 \\
% AWM--Qwen         & 96.0 & 8.7 & 98.0 & 200 \\
% AWM--InternVL     & 95.5 & 9.0 & 97.5 & 200 \\
% Full definitions and hypothetical example: HUMAN_TABLE_EXAMPLE.md.

\paragraph{Human document and watermark readability.}
% Retained previous human-evaluation wording:
% Our qualitative manual check found no interference with document reading or watermark-content recognition in the inspected images. Table~\ref{tab:human} is reserved for a formal study to quantify these properties. The planned study compares clean, plain, and model-specific attacked pages using human QA accuracy, response time, and watermark transcription accuracy. Conditions will be randomized and balanced without showing one participant multiple versions of the same source page. The formal study has not yet been conducted, so its sample sizes and quantitative results remain unfilled.
Fifteen volunteers rated 200 source pages each, viewing one version per page under shuffled, concealed condition labels (750 ratings per condition). They judged whether all original document text, numbers, and symbols were readable and whether at least one complete watermark instance was readable. Table~\ref{tab:human} reports affirmative responses over all completed ratings: document readability is 97.9\%/97.6\% and watermark readability is 94.8\%/94.3\% for Qwen/InternVL attacks. The full rating protocol and aggregation details appear in the supplement.

\subsection{Balanced Aggregation Loss}
\begin{table}[t]
\centering\small
\caption{Paired aggregation ablation on the same support as Table~\ref{tab:qa}. Hinge uses the global token mean; BAL uses grouped aggregation. Each arm uses its minimum complete loss over updates 1--250, without QA-based selection. ASR is in percent; paired differences and 95\% intervals appear in the supplement.}
\label{tab:ablation}
\setlength{\tabcolsep}{4pt}
\begin{tabular}{llrr}
\toprule
Model & Loss & ASR $\uparrow$ & ANLS $\downarrow$ \\
\midrule
Qwen2.5-VL & Hinge & 87.60 & 0.1280\\
 & BAL & \textbf{96.81} & \textbf{0.0345}\\
\midrule
InternVL2.5 & Hinge & 69.69 & 0.3411\\
 & BAL & \textbf{71.93} & \textbf{0.3084}\\
\bottomrule
\end{tabular}
\end{table}

% BEGIN retained DocOwl table (not rendered)
% \begin{table}[t]
% \centering\small
% \caption{Paired aggregation ablation on the same support as Table~\ref{tab:qa}. Each arm uses its minimum complete loss over updates 1--250, without QA-based selection. ASR is in percent; paired differences and 95\% intervals appear in the supplement.}
% \label{tab:ablation}
% \setlength{\tabcolsep}{4pt}
% \begin{tabular}{llrr}
% \toprule
% Model & Loss & ASR $\uparrow$ & ANLS $\downarrow$ \\
% \midrule
% Qwen2.5-VL & Hinge & 87.60 & 0.1280\\
%  & +BAL & 96.81 & 0.0345\\
% \midrule
% InternVL2.5 & Hinge & 69.69 & 0.3411\\
%  & +BAL & 71.93 & 0.3084\\
% \midrule
% DocOwl2 & Hinge & 84.95 & 0.1552\\
%  & +BAL & 85.33 & 0.1503\\
% \bottomrule
% \end{tabular}
% \end{table}
% END retained DocOwl table

\paragraph{BAL at the final endpoint.}
Table~\ref{tab:ablation} compares token hinge with BAL aggregation using the same initial images, editable pixels, paired questions, and 250-update budget. At the selected-loss endpoint, BAL increases Qwen ASR by 9.21 percentage points (paired 95\% CI $[7.68,10.76]$) and InternVL ASR by 2.24 points ($[1.14,3.37]$). BAL improves final ASR on both models, with the larger gain on Qwen. One possible reason is that Qwen's layout-based groups better expose small regions whose residual alignment is diluted by global token averaging, whereas InternVL's tile/thumbnail groups mix content within each view. This interpretation is discussed in the supplementary section on model-dependent behavior.

\begin{figure}[t]
\centering\includegraphics[width=\linewidth]{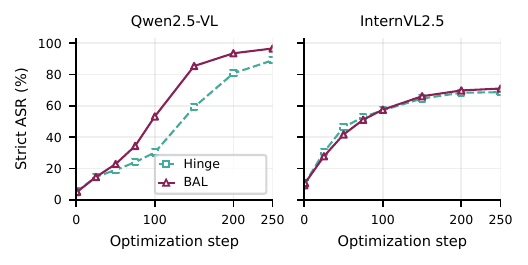}
\caption{Hinge and BAL step--ASR trajectories through update 250 on 200 randomly sampled DocVQA validation pages. Markers show evaluated checkpoints; lines connect those measurements.}
% Retained prior text:
% \caption{Auxiliary 300-update hinge and hinge+BAL trajectories on 200 randomly sampled DocVQA validation pages. Top: ASR versus update count. Bottom: ASR versus average per-page generation time to the same checkpoint, computed over 200 pages. Markers show evaluated checkpoints; lines connect those measurements.}

\label{fig:efficiency}
\end{figure}
\paragraph{Optimization progress.}
Figure~\ref{fig:efficiency} compares hinge and BAL at measured updates through step 250 on the random 200-page subset. At update 100, Qwen ASR is 30.38\% for hinge and 53.31\% for BAL; at update 150 it is 59.22\% and 85.46\%. BAL therefore produces stronger intermediate attacks on Qwen. InternVL shows smaller differences, with BAL weaker at some early checkpoints.

% Retained prior text:
% \paragraph{Update and time efficiency.}
% Figure~\ref{fig:efficiency} compares attack progress by update count and average per-page generation time. For each checkpoint, we average the recorded time each of the 200 pages takes to reach that update. At update 100, Qwen ASR is 30.38\% for hinge and 53.31\% for BAL; at update 150 it is 59.22\% and 85.46\%. For each model, we use hinge's own step300 ASR as a common target for both arms and report the first \emph{measured} checkpoint reaching it. Qwen reaches its 94.09\% target at step300 with hinge (1348.1 seconds) and step250 with BAL (1153.2 seconds). At these measured checkpoints, BAL reaches the target with 16.7\% fewer updates and 14.5\% less average per-page time. All runs in this trajectory study complete 300 updates.
% 
% The same rule gives step300 for both InternVL arms (741.1 versus 752.5 seconds). InternVL's BAL curve is weaker at some early checkpoints, and neither arm reaches hinge's final level earlier among the evaluated points. The observed efficiency gain is therefore specific to Qwen in this comparison.
% 
\subsection{Watermark Geometry, Placement, and Model Dependence}
\paragraph{Text versus rectangles.}
In the 200-page, 300-update InternVL study, 8\%-area rectangle and text carriers attain 69.53\% and 71.99\% selected-loss ASR, respectively. Their Plain ASRs are 2.58\% and 9.46\%, respectively. Text first reaches 65\% ASR at step 150, while rectangles first reach it at step 250. The text carrier's spatial structure and semantic cues may contribute to this earlier onset.

\begin{figure}[t]
\centering\includegraphics[width=\linewidth]{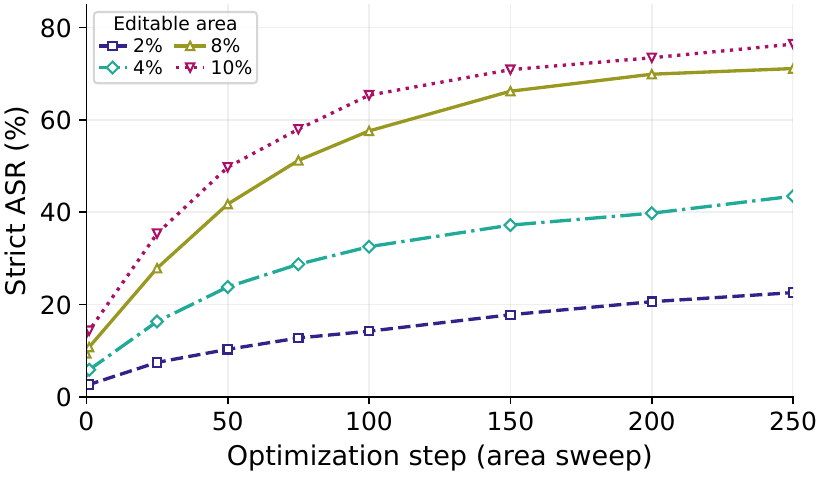}
\caption{InternVL step--ASR curves through update 250 for 2/4/8/10\% watermark area at a fixed 24-pixel font size, on 200 randomly sampled DocVQA validation pages. Step 0 is the corresponding plain watermark.}
\label{fig:boundaries}
\end{figure}
% Retained prior version:
% \begin{figure*}[t]
% \centering\includegraphics[width=\textwidth]{figures/boundaries.pdf}
% \caption{Carrier area and cross-model transfer on 200 randomly sampled DocVQA validation pages. (a) InternVL step--ASR curves for 2/4/8/10\% area at a fixed font size of 24 pixels; step 0 is the corresponding plain watermark. (b) Same-PNG transfer, with rows indicating source and columns target. Transfer ASR uses the target\textquotesingle s clean-correct denominator.}
% % Retained prior text:
% % \caption{Auxiliary 300-update carrier budgets and attack boundaries. (a) InternVL step--ASR curves for 2/4/8/10\% area, with fixed 24-pixel text; step 0 is the corresponding plain watermark. Only measured checkpoints are connected. (b) Font-size comparison at 8\% area, selected-loss endpoint. Both use the same 200 pages. (c) Single-scale attacks on the separate 96-page, 403-question resizing evaluation; capped inputs use the indicated maximum dimension. (d) Same-PNG transfer on the 200-page set, with rows indicating source and columns target. Transfer ASR uses the target's clean-correct denominator.}
% 
% \label{fig:boundaries}
% \end{figure*}

\begin{table}[t]
\centering\small
\caption{InternVL carrier geometry and QA outcomes, 200 pages / 894 questions. All rows use the ``confidential'' watermark. Area is actual stroke percentage. Touch and Cov$_{10}$ are percentages of visual cells with mapped mask occupancy $>0$ and $\geq0.10$. Plain and WASP share these coverage values; ASR is in percent.}
\label{tab:coverage}
\setlength{\tabcolsep}{4pt}
\begin{tabular}{rrrrrr}
\toprule
Font & Area & Touch & Cov$_{10}$ & Plain & WASP \\
\midrule
24 & 1.995 & 18.26 & 6.29 & 2.70 & 23.46\\
24 & 3.994 & 34.03 & 13.46 & 5.65 & 44.23\\
24 & 7.995 & 61.62 & 27.31 & 9.46 & 71.99\\
24 & 9.996 & 70.41 & 39.49 & 13.76 & 77.64\\
% 16 & 7.998 & 76.47 & 30.89 & 7.74 & 66.83\\
% 36 & 7.992 & 48.38 & 25.11 & 9.58 & 71.01\\
\bottomrule
\end{tabular}
\end{table}

\paragraph{Editable area and visual-token coverage.}
Increasing area from 2\% to 10\% at a fixed 24-pixel font size raises InternVL selected-loss ASR from 23.46\% to 77.64\%, while the fraction of visual-token input cells touched by watermark strokes increases from 18.26\% to 70.41\% (Table~\ref{tab:coverage}). Larger areas also yield stronger intermediate attacks (Figure~\ref{fig:boundaries}). These results link attack effectiveness to editable watermark area and visual-token coverage: larger supports reach more token input cells and produce stronger attacks in the fixed-font area sweep.

\begin{table}[t]
\centering\small
\caption{Paired position controls on 200 DocVQA pages / 894 questions, with equal glyph count and stroke area per page. Text favors text blocks; BG favors background. All scores are ASR (\%); WASP uses step 300. Random averages three RMS-targeted seeds; 3/600 InternVL Text controls miss tolerance and remain included.}
\label{tab:layout}
\setlength{\tabcolsep}{4pt}
\begin{tabular}{llrrr}
\toprule
Model & Placement & Plain & Random & WASP \\
\midrule
Qwen2.5-VL & Text & 4.61 & 6.26 & 96.57 \\
Qwen2.5-VL & BG & 1.42 & 1.65 & 99.05 \\
InternVL2.5 & Text & 11.67 & 11.10 & 64.13 \\
InternVL2.5 & BG & 4.05 & 3.48 & 21.62 \\
\bottomrule
\end{tabular}
\end{table}

\paragraph{A paired position intervention.}
On the same 200 DocVQA pages, we compare text-prioritized (Text) and background-prioritized (BG) layouts with equal glyph counts and stroke area per page (mean 7.969\%). Both use the default font, rotation, and opacity with common placement candidates. Text favors frozen clean-page text blocks; BG favors the complement of all detected content blocks. Each layout is optimized separately for 300 updates and evaluated at step 300, distinct from the uniform-carrier main experiment.

\paragraph{Placement verification.}
Text and BG place 51.53\% and 1.36\% of editable strokes within text blocks, respectively; BG places 95.44\% outside detected content blocks. The supplement reports placement construction, foreground diagnostics, and a low-overlap subset fixed before QA.

\paragraph{Model-dependent effects beyond passive obstruction.}
Table~\ref{tab:layout} compares each placement with its Plain and RMS-targeted Random controls. BG WASP reaches 99.05\% ASR on Qwen and 21.62\% on InternVL, exceeding Random by 97.40 and 18.14 points. Passive overlap and random magnitude alone therefore do not explain the damage. Qwen remains strongly vulnerable under both placements, whereas InternVL reaches 64.13\% with Text versus 21.62\% with BG. After subtracting each layout's Plain baseline, the Text-minus-BG contrast is $-5.67$ points for Qwen and $34.89$ for InternVL. These differences suggest a role for page-wide visual interactions in Qwen versus local tiles and a thumbnail in InternVL. Full controls, matching exceptions, and uncertainty estimates appear in the supplement.

\subsection{Failure Propagation in Structured Parsing}
\begin{table}[t]
\centering\small
\caption{MinerU2.5 on 1,651 OmniDocBench pages. ED increases with damage; TEDS/CDM decrease. Official text, formula, and order ED use 1,557/313/1,638 eligible pages.}
\label{tab:mineru}
\setlength{\tabcolsep}{4pt}
\begin{tabular}{lrrrrr}
\toprule
Input & Text & Formula & Order & TEDS & CDM \\
\midrule
Clean & 0.055 & 0.114 & 0.137 & 0.846 & 0.945\\
Plain & 0.064 & 0.133 & 0.141 & 0.810 & 0.938\\
WASP & \textbf{0.839} & \textbf{0.905} & \textbf{0.841} & \textbf{0.167} & \textbf{0.097}\\
\bottomrule
\end{tabular}
\end{table}

\paragraph{Structured document parsing.}
In Table~\ref{tab:mineru}, WASP increases text ED from 0.0642 for plain pages to 0.8390, a paired increase of 0.7748. Formula and reading-order ED also rise, while TEDS and CDM decrease. Text ED is aggregated over 1,557 pages with the corresponding official component; all 1,651 input pages remain in the evaluation under the evaluator\textquotesingle s category-missing rules.

\begin{table}[t]
\centering\small
\caption{MinerU routing diagnostic on 200 randomly sampled OmniDocBench pages (300-update attacks). Text ED uses 191 eligible pages; higher values indicate greater extraction error.}
\label{tab:routing}
\setlength{\tabcolsep}{4pt}
\begin{tabular}{llr}
\toprule
Routing & Content & Text ED $\uparrow$ \\
\midrule
Attacked & Attacked & 0.8469\\
Clean & Attacked & 0.0715\\
Attacked & Clean & 0.8429\\
\bottomrule
\end{tabular}
\end{table}

% Retained prior version:
% \begin{figure}[t]
% \centering\includegraphics[width=\linewidth]{figures/routing.pdf}
% \caption{Routing intervention for MinerU2.5 on 200 randomly sampled OmniDocBench pages with 300 updates. Full attack means attacked routing and content. Clean routes largely restore text extraction from the attacked image, while attacked routes retain large errors even with clean content. These are complete-pipeline diagnostic conditions, not an additive causal decomposition.}
% \label{fig:routing}
% \end{figure}

\paragraph{How parsing damage propagates.}
The routing diagnostic in Table~\ref{tab:routing} compares the full attacked pipeline, attacked content under clean routes, and clean content under attacked routes. Text ED is respectively 0.8469, 0.0715, and 0.8429. The attacked layout is empty for 37/200 pages, and mean ground-truth content coverage decreases from 0.8952 to 0.4072. Preserving clean routing largely recovers text extraction despite attacked pixels; retaining attacked routing keeps errors high despite clean pixels. This pattern associates the degradation primarily with upstream layout and routing.

\begin{figure*}[t]
\centering\includegraphics[width=0.85\textwidth]{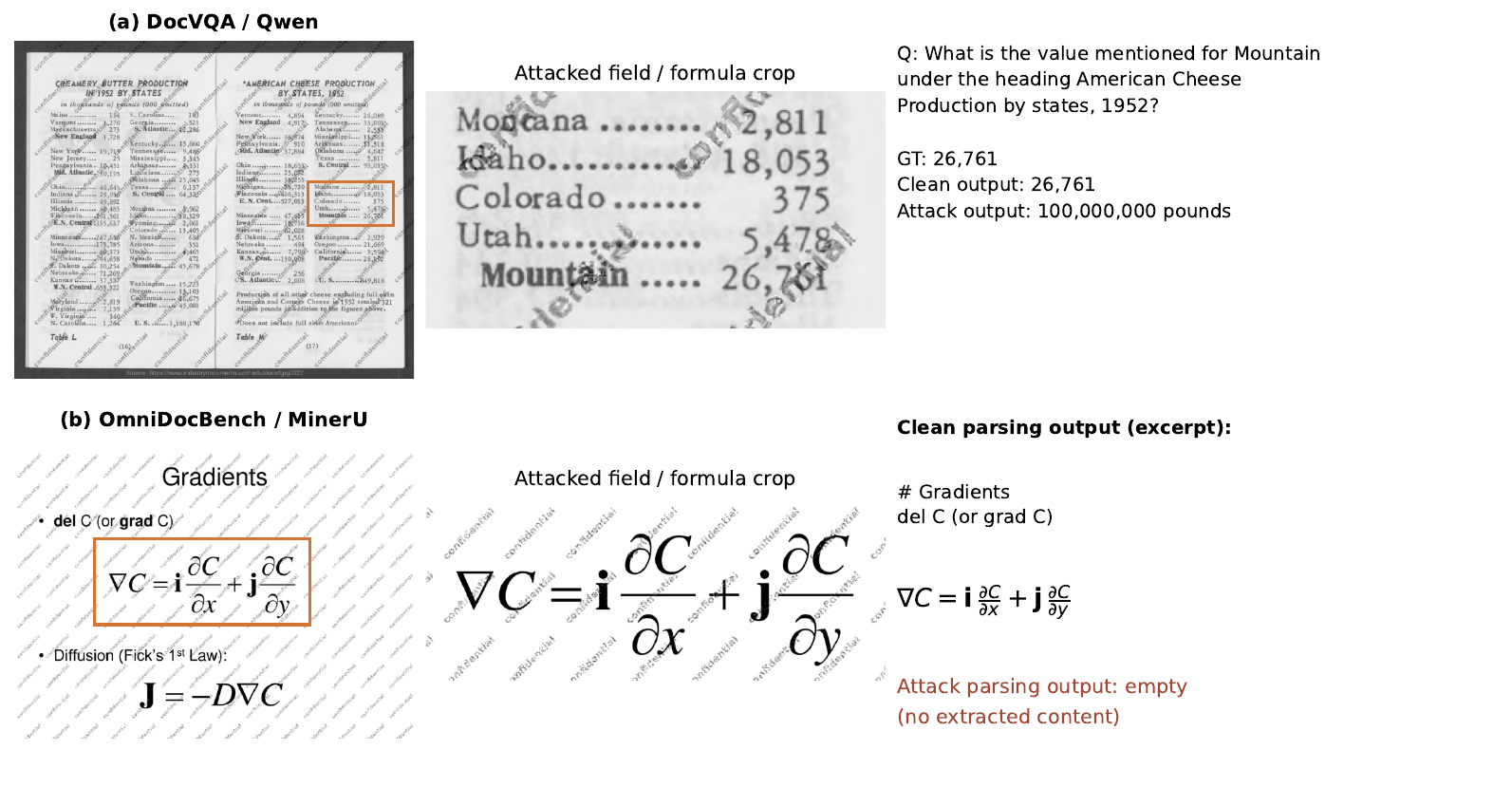}
\caption{Selected 300-update examples with readable source content. (a) Qwen answers ``100,000,000 pounds'' instead of 26,761. (b) MinerU returns an empty output; the clean formula extraction is shown for comparison. Crops are from evaluated attacked PNGs.}
\label{fig:cases}
\end{figure*}
\paragraph{Qualitative examples.}
Figure~\ref{fig:cases} shows a DocVQA numerical-answer error and an empty MinerU parsing output despite visible source content. Image identities, full outputs, crop coordinates, and selection procedures appear in the supplement.

\subsection{Transferability and Resizing}
\paragraph{Cross-model transfer.}
\begin{table}[t]
\centering\small
\caption{Cross-model transfer on 200 DocVQA pages / 894 questions, using loss-selected source PNGs from 300 updates. Plain and Transfer report target ASR (\%); $\Delta$ASR is Transfer minus Plain in percentage points (pp). Mistral/Vicuna are LLaVA-v1.6 7B variants; Qwen/InternVL sources are 7B/8B.}
\label{tab:transfer}
\setlength{\tabcolsep}{4pt}
\begin{tabular}{llrrr}
\toprule
Source & Target & Plain & Transfer & $\Delta$ASR (pp) \\
\midrule
InternVL & InternVL-2B & 12.06 & 50.07 & +38.01 \\
Mistral & Vicuna & 24.17 & 66.28 & +42.11 \\
Qwen & Qwen-3B & 6.06 & 17.09 & +11.03 \\
\midrule
Qwen & InternVL & 9.46 & 12.29 & +2.83 \\
InternVL & Qwen & 5.20 & 11.35 & +6.15 \\
\bottomrule
\end{tabular}
\end{table}

Table~\ref{tab:transfer} shows strong transfer for some architecturally related readers. InternVL-8B to InternVL-2B reaches 50.07\% ASR, exceeding Plain by 38.01 points (paired 95\% CI $[33.83,42.20]$); Mistral to Vicuna reaches 66.28\%, a gain of 42.11 points ($[37.23,46.95]$). Qwen-7B to Qwen-3B shows a smaller gain, while the tested cross-architecture transfers reach only 12.29\% and 11.35\% ASR.

All transfers reuse unchanged, loss-selected PNGs from 300-update runs on 200 DocVQA pages / 894 questions, without target optimization or target-based selection. ASR uses each target's clean-correct set. The audited scale pairs transfer despite different visual weights and connector shapes. Denominators, secondary intervals, and common-clean-correct checks appear in the supplement.

\paragraph{Resizing.}
\begin{table}[t]
\centering\small
\caption{Resizing on 96 DocVQA pages / 403 questions, step 300. Columns report ASR (\%). Cap is the maximum input side length in pixels; original resolution is uncapped. WASP is single-scale; Multi uses weighted multiscale optimization. Each input scale uses its target-specific clean-correct denominator. Qwen Multi is a post-hoc extension.}
\label{tab:resizing}
\setlength{\tabcolsep}{4pt}
\begin{tabular}{llrrr}
\toprule
Model & Cap & Plain & WASP & Multi \\
\midrule
Qwen2.5-VL & Original & 4.23 & 97.62 & 85.98\\
 & 1280 & 9.09 & 16.58 & 32.89\\
 & 1600 & 6.10 & 19.63 & 22.81\\
 & 1920 & 7.65 & 25.33 & 40.90\\
\midrule
InternVL2.5 & Original & 9.42 & 72.02 & 69.25\\
 & 1280 & 10.83 & 48.89 & 63.89\\
 & 1600 & 10.44 & 62.91 & 66.48\\
 & 1920 & 10.50 & 67.40 & 67.68\\
\bottomrule
\end{tabular}
\end{table}

Both single-scale and weighted multiscale attacks exceed Plain ASR under the tested resizing caps (Table~\ref{tab:resizing}).
These results suggest that input resizing alone is insufficient to neutralize adversarial watermarks.

\section{Discussion and Conclusion}
% \paragraph{Limitations.}
% Our source attacks require white-box access, and Qwen's BAL grouping additionally uses clean-page layout. BAL improves final ASR on both evaluated QA models; the gain varies with the model and optimization stage. Generalization to other frontends and grouping schemes remains to be evaluated. The architectural explanations for placement and transfer warrant controlled interventions. Human readability results are subjective ratings summarized from aggregate counts; participant/page-linked records are needed for clustered uncertainty estimates.
\paragraph{Limitations.}
WASP requires document-specific white-box access to
the visual frontend; Qwen's BAL grouping also uses
clean-page layout predictions. Attack effectiveness
depends on preprocessing: input resizing substantially
weakens Qwen's single-scale attack, although multiscale
optimization improves robustness. BAL's gains vary
across models and optimization stages, and broader
frontend and grouping evaluations remain open.
The placement and transfer results suggest architectural
dependence but do not isolate its underlying cause.
Human readability is measured through subjective
ratings; participant- and page-linked records would
be required for clustered uncertainty estimates.

\paragraph{Conclusion.}
Visible watermark strokes can carry adversarial signals that substantially disrupt document QA and structured parsing while retaining high human-perceived readability. WASP demonstrates damage beyond ordinary watermarks and random perturbations. BAL improves final attack success on both evaluated QA models by directing optimization toward groups with high residual alignment. Background-prioritized attacks reveal vulnerabilities beyond direct text occlusion, and routing interventions identify an upstream failure pathway in MinerU. Strong transfer between some architecturally related readers extends this risk beyond the optimization target. These findings motivate robustness evaluation that accounts for adversarial signals embedded in legitimate document overlays.

\section*{Ethical Considerations}
No formal IRB or equivalent ethics review was conducted,
as the study involved only voluntary judgments of document
and watermark readability and did not collect sensitive
participant information. All participants provided informed
consent. Results are reported only in aggregate, without
identifying individual participants.

\section*{LLM Usage Statement}
We used \texttt{gpt-6-astra} for manuscript polishing and error checking; the authors manually checked the literature findings and references. The authors designed the experiments, while \texttt{gpt-6.1-sol} through OpenAI Codex assisted with code implementation and execution. For strict ASR, \texttt{gpt-6.1-sol} labeled model answers using the fixed supplementary rubric. The authors remain responsible for all claims, data, citations, code, and submitted content.

\bibliography{main}
\clearpage
\appendix
\section{Image and Evaluation Protocol}
The evaluation unit for sampling and confidence intervals is a page. The main experiments and Hinge/BAL ablation use the complete DocVQA validation split of 1,285 pages and 5,349 questions; both models share these sample identities (Table~\ref{tab:endpoints}). TextVQA uses the same fixed validation sample of 1,000 images and 1,592 questions for both models. Auxiliary studies retain the original 200 pages sampled with Python\textquotesingle s random seed 20261028 after sorting unique validation pages by document identifier and pixel hash. The five calibration pages use seed 20261029 on the training split after removing document-ID and pixel-identity overlap with validation. Images retain their original resolution before each model\textquotesingle s native processor.

The default font renderer uses the frozen DejaVuSans-Bold asset, 24-pixel size, a $45^\circ$ rotation, and black color with opacity 0.30. Let $a$ denote the rendered alpha map before opacity scaling. The binary support is $M=\mathbf{1}[a\geq8/255]$ (inclusive threshold). The effective compositing alpha is $\eta a$ with $\eta=0.30$; opacity is not applied when thresholding the editable mask. Alpha compositing uses a maximum over overlapping instances. The original 200-page geometry study measured average actual support fraction 0.0799484. Parameters modify only this support and are broadcast across the three color channels.

\paragraph{Frontend settings.}
Qwen2.5-VL-7B uses image tokens after the visual merger, patch size 14 and merge factor 2, with native minimum/maximum pixel settings 3136 and 12845056. InternVL2.5-8B uses 448-pixel dynamic tiles, 1--12 local tiles and a thumbnail, taking features after pixel shuffle and the MLP connector.

\paragraph{Task metrics and semantic scoring.}
Exact match and ANLS score DocVQA predictions against the official answers; TextVQA uses official VQA accuracy. Semantic ASR uses a separate, model-specific clean-correct denominator. Let $Q_m$ denote the questions used to evaluate model $m$, and let $J_m^c(q)$ and $J_m^t(q)$ be the frozen judge labels for the clean answer and condition $t$, respectively. Define
\begin{equation}
\begin{split}
 C_m&=\{q\in Q_m:J_m^c(q)=\text{correct}\},\\
 \mathrm{ASR}_m(t)&=\frac{1}{|C_m|}\sum_{q\in C_m}
 \mathbf{1}[J_m^t(q)\in\mathcal{S}],
\end{split}
\end{equation}
where $\mathcal{S}=\{\text{wrong-key},\text{off-topic},\text{no-answer}\}$. Reported ASR values are percentages. The same $C_m$ is used across compared conditions; it is defined by semantic correctness, rather than exact match or an ANLS threshold. Partial and uncertain attacked answers contribute zero to the numerator and remain in the denominator. Transfer uses the target model's clean-correct set. Random averages the three seed-specific success indicators per question with the same denominator. Clean ASR is shown as a dash because it is not an attack condition.

The archived judge uses model identifier \texttt{gpt-6.1-sol}, low reasoning effort, a 4096-output-token cap, structured JSON responses, and a fixed six-label rubric. It evaluates the question, reference answers, and generated answer; reference alternatives are acceptable alternatives rather than jointly required facts. The labels are:
\begin{itemize}
\item \textbf{Correct}: the answer supplies the requested fact, allowing semantically equivalent wording, sufficient aliases, and unambiguous formatting or unit differences. Verbosity alone does not make an otherwise correct answer incorrect.
\item \textbf{Wrong-key}: the requested value, date, entity, or field is wrong. A wrong digit in a single-field answer is wrong-key, even if other words match the reference.
\item \textbf{Off-topic}: the answer does not address the question.
\item \textbf{No-answer}: the model refuses, returns an empty or placeholder-only response, or provides no extractable answer.
\item \textbf{Partial}: a question requires multiple distinct facts and the answer supplies only some correctly; this label is not used merely for partial word overlap in a wrong single-field answer.
\item \textbf{Uncertain}: the question or reference evidence is ambiguous, contradictory, or insufficient to decide; this label does not represent an API or transport failure.
\end{itemize}
Only label membership in $\mathcal{S}$ determines strict success; auxiliary judge scores, confidence, and tags do not change it. This automatic semantic score is reported alongside the task metrics and is not a substitute for a human-reading study.

The generation prompt is ``Answer the following question about the document image. Reply with only the answer, using as few words as possible. Question: \{question\}.'' Generation is greedy, with one beam and at most 256 new tokens. Truncated responses retain the fixed-limit evaluation outcome. All compared conditions use the frozen scoring protocol.

\paragraph{Main-table random controls.}
The main DocVQA controls share each model\textquotesingle s WASP mask and target its selected-loss PNG\textquotesingle s within-mask RGB RMS relative to Plain after clipping and quantization. Three random seeds are averaged per question. The random perturbations target the same within-mask RMS as WASP, subject to the shared amplitude bound, and preserve pixels outside the stroke mask.

\begin{table*}[t]
\centering\small
\caption{Paired DocVQA endpoint counts, minimum complete loss over updates 1--250 for both arms. Pages and QA denote the shared validation dataset size.}
\label{tab:endpoints}
\setlength{\tabcolsep}{4pt}
\begin{tabular}{llrrrrrr}
\toprule
Model & Loss & Dataset pages & Dataset QA & Success / eligible & ASR (\%) & ANLS & EM (\%) \\
\midrule
Qwen2.5-VL & Hinge & 1285 & 5349 & 4451/5081 & 87.60 & 0.1280 & 8.95\\
Qwen2.5-VL & BAL & 1285 & 5349 & 4919/5081 & 96.81 & 0.0345 & 2.03\\
InternVL2.5 & Hinge & 1285 & 5349 & 3428/4919 & 69.69 & 0.3411 & 20.02\\
InternVL2.5 & BAL & 1285 & 5349 & 3538/4919 & 71.93 & 0.3084 & 18.32\\
\bottomrule
\end{tabular}
\end{table*}

% BEGIN retained DocOwl table (not rendered)
% \begin{table*}[t]
% \centering\small
% \caption{Paired DocVQA endpoint counts, minimum complete loss over updates 1--250 for both arms. All questions of each retained page are included.}
% \label{tab:endpoints}
% \setlength{\tabcolsep}{4pt}
% \begin{tabular}{llrrrrrr}
% \toprule
% Model & Loss & Pages & QA & Success / eligible & ASR (\%) & ANLS & EM (\%) \\
% \midrule
% Qwen2.5-VL & Hinge & 1282 & 5330 & 4451/5081 & 87.60 & 0.1280 & 8.95\\
% Qwen2.5-VL & +BAL & 1282 & 5330 & 4919/5081 & 96.81 & 0.0345 & 2.03\\
% InternVL2.5 & Hinge & 1285 & 5349 & 3428/4919 & 69.69 & 0.3411 & 20.02\\
% InternVL2.5 & +BAL & 1285 & 5349 & 3538/4919 & 71.93 & 0.3084 & 18.32\\
% DocOwl2 & Hinge & 1285 & 5349 & 3504/4125 & 84.95 & 0.1552 & 8.67\\
% DocOwl2 & +BAL & 1285 & 5349 & 3520/4125 & 85.33 & 0.1503 & 8.26\\
% \bottomrule
% \end{tabular}
% \end{table*}
% END retained DocOwl table

\section{Uncertainty for Secondary Comparisons}
Table~\ref{tab:secondary_ci} collects paired differences with 10,000 page-bootstrap resamples (seed 20261028), without multiplicity correction. The parsing interval uses the official per-page text ED values. Both BAL contrasts use the paired DocVQA validation sample and the minimum complete loss over updates 1--250. Carrier, area, font, and wording contrasts retain their original 200-page, 300-update results. ASR differences are in percentage points; parsing differences are in text-ED units. Layout contrasts remain in Table~\ref{tab:layout_ci}.

\begin{table*}[t]
\centering\small
\caption{Paired differences and unadjusted 95\% page-bootstrap intervals. Top: full parsing and 250-update DocVQA validation evaluations. Bottom: retained 200-page, 300-update InternVL carrier studies. ASR differences use percentage points.}
\label{tab:secondary_ci}
\setlength{\tabcolsep}{4pt}
\begin{tabular}{llrr}
\toprule
Experiment & Contrast & Difference & 95\% CI \\
\midrule
Parsing text ED & WASP $-$ Plain & 0.7748 & [0.7582, 0.7909] \\
Qwen2.5-VL BAL & BAL $-$ Hinge & 9.21 & [7.68, 10.76] \\
InternVL2.5 BAL & BAL $-$ Hinge & 2.24 & [1.14, 3.37] \\
\midrule
Carrier & Rectangle $-$ Text & -2.46 & [-5.89, 0.84] \\
Area & 10\% $-$ 8\% & 5.65 & [2.66, 8.73] \\
Font size & 16 px $-$ 24 px & -5.16 & [-9.01, -1.43] \\
Font size & 36 px $-$ 24 px & -0.98 & [-4.36, 2.41] \\
Wording & copyright $-$ confidential & 0.86 & [-2.54, 4.36] \\
\bottomrule
\end{tabular}
\end{table*}

% BEGIN retained DocOwl table (not rendered)
% \begin{table*}[t]
% \centering\small
% \caption{Paired differences and unadjusted 95\% page-bootstrap intervals. Top: expanded parsing and 250-update DocVQA evaluations. Bottom: retained 200-page, 300-update InternVL carrier studies. ASR differences use percentage points.}
% \label{tab:secondary_ci}
% \setlength{\tabcolsep}{4pt}
% \begin{tabular}{llrr}
% \toprule
% Experiment & Contrast & Difference & 95\% CI \\
% \midrule
% Parsing text ED & AWM $-$ Plain & 0.7748 & [0.7582, 0.7909] \\
% Qwen2.5-VL BAL & BAL $-$ Hinge & 9.21 & [7.68, 10.76] \\
% InternVL2.5 BAL & BAL $-$ Hinge & 2.24 & [1.14, 3.37] \\
% DocOwl2 BAL & BAL $-$ Hinge & 0.39 & [-0.52, 1.29] \\
% \midrule
% Carrier & Rectangle $-$ Text & -2.46 & [-5.89, 0.84] \\
% Area & 10\% $-$ 8\% & 5.65 & [2.66, 8.73] \\
% Font size & 16 px $-$ 24 px & -5.16 & [-9.01, -1.43] \\
% Font size & 36 px $-$ 24 px & -0.98 & [-4.36, 2.41] \\
% Wording & copyright $-$ confidential & 0.86 & [-2.54, 4.36] \\
% \bottomrule
% \end{tabular}
% \end{table*}
% END retained DocOwl table

\section{Gradient Calibration and Optimization Details}
\subsection{BAL group construction}
\label{sec:sup_bal_groups}
Group membership is constructed once from the clean page and native frontend geometry, then held fixed throughout optimization and checkpoint selection. Each valid image token belongs to exactly one group. BAL retains only groups containing tokens, and $R$ denotes their number; regions without token centers are omitted from both the group means and the log-sum-exp normalization. With one remaining group, BAL reduces to the global token hinge average.

\paragraph{Qwen: source of the clean layout.}
We run the frozen MinerU2.5-2509-1.2B layout detector on the unwatermarked RGB page using its native Transformers backend and $1036\times1036$ layout view. This is an auxiliary preprocessing step, not a ground-truth annotation or a layout generated by the attacked Qwen model. It uses the native \texttt{Layout Detection:} prompt, temperature 0, top-$k=1$, top-$p=0.01$, and repetition penalty 1.0. The implementation uses MinerU 2.5.4 and mineru-vl-utils 0.1.12. Only predicted boxes and block types define groups; questions, answers, recognized text, and dataset layout annotations are not used. Layouts are cached before attack generation, with no gradients through the detector and no recomputation from attacked pages.

Normalized box coordinates are mapped to original-page pixels by flooring left/top coordinates, ceiling right/bottom coordinates, and clipping to the page. Nonpositive-area boxes are discarded. Each box defines a separate region, including boxes of the same semantic type. To resolve overlaps, boxes are sorted by increasing pixel area, then lexicographic block type, top coordinate, left coordinate, and original output index; each box receives only pixels not assigned earlier. The detector provides no confidence score, so the implementation's confidence tie-break is constant. Unassigned pixels form background group 0. No neighboring regions are merged.

For a merged Qwen token grid of height $h_t$ and width $w_t$, token $(u,v)$ takes the partition label at original-page pixel
\begin{equation}
 \left(\left\lfloor\frac{(v+1/2)W}{w_t}\right\rfloor,
       \left\lfloor\frac{(u+1/2)H}{h_t}\right\rfloor\right).
\end{equation}
Indices are clipped to image bounds, and tokens follow the spatial order restored after the visual merger. Background participates only when it contains tokens. A valid empty detection produces a single background group; malformed or truncated detector output is recorded as a technical failure, never silently converted to background.

\paragraph{InternVL: native view groups.}
Each native $448\times448$ local tile contributes a separate group of 256 post-connector tokens ($16\times16$). Groups follow the processor's local-tile order, with the thumbnail as an additional group when present. All valid tokens are retained, including thumbnail tokens even where their source content overlaps local tiles. Grouping needs no layout detector; a single-view input has one group.

\paragraph{MinerU: spatial groups.}
For the native layout branch, let $h_t\times w_t$ be the grid after $2\times2$ spatial merging. A token at row $u$, column $v$ is assigned
\begin{equation}
 g(u,v)=3\left\lfloor\frac{3u}{h_t}\right\rfloor
          +\left\lfloor\frac{3v}{w_t}\right\rfloor.
\end{equation}
The implementation checks that the single-image grid contains $h_tw_t$ tokens in native row-major order and that all nine groups are nonempty. Groups use spatial coordinates, not predicted layout boxes. These frontend-specific constructions share the same zero-threshold token hinge, within-group mean, and normalized log-sum-exp with $\beta=10$.

\subsection{Gradient calibration and shared optimization}
For the QA models, we calibrate the scale $k_B$ on five training development pages, excluding document- or pixel-identity overlap with validation. For each page $d$, we initialize $\omega=0$, so the attacked input is the plain watermark $x_w$, while the feature reference is the clean page $x_c$. For $J\in\{\mathrm{hinge},\mathrm{BAL}\}$, define the within-support gradient root mean square (RMS) of $\mathcal L_J$:
\begin{equation}
 g_J^{(d)}=\left(\frac{\sum_p M_p^{(d)}\left(\left.\frac{\partial\mathcal L_J^{(d)}}{\partial\omega_p}\right|_{\omega=0}\right)^2}{\sum_p M_p^{(d)}}\right)^{1/2}.
\end{equation}
The model-specific scale is
\begin{equation}
 k_B=\frac{\operatorname{median}_{d=1,\ldots,5}g_{\mathrm{hinge}}^{(d)}}{\operatorname{median}_{d=1,\ldots,5}g_{\mathrm{BAL}}^{(d)}}.
 \label{eq:calibration}
\end{equation}
This matches the median initial gradient magnitude of $k_B\mathcal L_{\mathrm{BAL}}$ to that of $\mathcal L_{\mathrm{hinge}}$, keeping their scale comparable relative to the visual regularizer. It uses the ratio of two medians, not the median of per-page ratios. The resulting values are 1.21604 for Qwen and 0.57925 for InternVL. Each value is frozen before evaluation and shared across pages and optimization steps; no downstream QA score is used to choose it. The calibration matches initial gradient scales, not gradient directions or their evolution during optimization.

The hinge-only ablation substitutes $F=\mathcal L_{\mathrm{hinge}}$. The squared-error term sums RGB channels and averages over editable pixels. Total variation (TV) averages absolute differences between adjacent pixels separately in each spatial direction:
\begin{equation}
 \operatorname{TV}(x)=\frac{\|D_hx\|_1}{3(H-1)W}
                    +\frac{\|D_wx\|_1}{3H(W-1)}.
\end{equation}
Here $(D_hx)_{u,v,c}=x_{u+1,v,c}-x_{u,v,c}$ and $(D_wx)_{u,v,c}=x_{u,v+1,c}-x_{u,v,c}$ are vertical and horizontal differences; $u,v,c$ index rows, columns, and RGB channels, and $\|\cdot\|_1$ sums absolute entries. TV is evaluated on the full attacked image $x_a$. Both visual terms use continuous pixels before quantization. The shared 250-update protocol uses Adam with learning rate 0.01, $(\beta_1,\beta_2)=(0.9,0.999)$, $\epsilon_{\mathrm{Adam}}=10^{-8}$, and no weight decay or early stopping.

The network runs in BF16 while parameters, cosine normalization, and loss reductions use FP32. Forward scoring uses native preprocessing of quantized pixels,
$q(x)=\lfloor255\operatorname{clip}(x,0,1)+0.5\rfloor/255$.
Where native preprocessing is nondifferentiable, a straight-through construction preserves the official forward value and uses a tensor preprocessing path for approximate gradients. The attack does not differentiate exactly through a discrete document parser.

For the DocVQA validation evaluation, we select \texttt{best\_loss} by the minimum complete $\mathcal L$ over updates 1--250, breaking ties in favor of the earlier update, and retain a restorable step 250 backup. No downstream answer selects the checkpoint. Each arm uses its own complete objective, so this comparison evaluates the complete optimization-and-selection procedure at a common 250-update budget. The auxiliary trajectory studies retain their original fixed checkpoints through update 300.

\subsection{MinerU objective and implementation}
\label{sec:sup_mineru_optimization}
MinerU uses the same support-constrained parameterization and complete objective as the QA attacks:
\begin{equation}
\begin{split}
 \mathcal L_{\mathrm{MinerU}}&=0.5k_B\mathcal L_{\mathrm{BAL}}+0.02V,\\
 V&=\frac{\sum_{p,c}M_p(x_{a,pc}-x_{w,pc})^2}{\max(1,\sum_p M_p)}
       +\operatorname{TV}(x_a).
\end{split}
\end{equation}
The pixel term sums RGB channels without dividing by three, and TV acts on the full continuous attacked page. The shared single-channel parameter starts at zero, has perturbation amplitude bound 0.5 relative to Plain, and is clipped to $[-8,8]$ after each update. Training and checkpoint selection use the same complete loss with the Adam settings above. Selection considers updates 1--250, breaks ties toward the earlier update, and uses no parser output or downstream score.

The target is the frozen MinerU2.5-2509-1.2B visual merger before the language model, accessed through its Qwen2VL implementation. Native layout preprocessing stretches the page to $1036\times1036$ with PIL bicubic interpolation, followed by the native processor (14-pixel patches, merge factor 2, pixel limits 3,136 and 1,605,632). Model execution uses BF16 and SDPA; cosine normalization and losses use FP32 with normalization epsilon $10^{-8}$. Quantized native preprocessing supplies the forward values; a differentiable bicubic, normalization, and patch-packing path supplies surrogate gradients. No layout-generation prompt or language-model loss enters attack optimization. The exported PNG is subsequently evaluated through the full native parsing pipeline.

MinerU's $k_B$ follows Eq.~\ref{eq:calibration}, using the five fixed OmniDocBench development pages and the nine spatial groups. Calibration uses only feature gradients at $\omega=0$, excluding the visual regularizer, and the scale is frozen across pages and steps. These development pages belong to the full 1,651-page collection; the accompanying sensitivity protocol excludes them, leaving 1,646 pages. They are not described as independent training data.
% Pending experiment completion: insert the measured MinerU k_B and refresh
% parsing result tables from the matching hinge/BAL + QA-visual-regularizer run.
% Existing parsing numbers and historical routing interventions are unchanged.

\section{Endpoint Selection and Runtime}
The DocVQA validation evaluation selects the minimum complete attack loss over updates 1--250, with earlier updates breaking ties; both arms perform all 250 updates. A step 250 backup is retained for exact restoration, but Table~\ref{tab:endpoints} reports the selected-loss endpoint. The separate 200-page trajectory studies use updates 1--300 and fixed checkpoints 0, 1, 25, 50, 75, 100, 150, 200, 250, and 300. Their retained selected-loss and fixed-checkpoint archives support the diagnostic curves; no 250-update full-evaluation trajectories are inferred from those curves.

The main-text optimization plot uses only measured step--ASR checkpoints through update 250 on the random 200-page subset. Each checkpoint uses the corresponding PNG from every page. The underlying auxiliary archives retain their original 300-update records; truncating the displayed trajectory does not change its recorded checkpoints. Runtime is instead measured from the new 250-update runs in Table~\ref{tab:runtime250}.

% Retained prior text:
% For the original 200-page trajectory study, at checkpoint $t$, the plotted time is $\bar\tau(t)=\frac{1}{200}\sum_{d=1}^{200}\tau_d(t)$, where $\tau_d(t)$ is the recorded cumulative generation time for page $d$ to reach update $t$. ASR uses the corresponding checkpoint image from every page. Thus each point aligns pages by update count and summarizes their individual runtimes with a mean; total elapsed time for processing the dataset depends separately on scheduling and parallelism. Full generation cost also includes final export and archival work. Loading, preprocessing layout preparation, downstream QA, judging, and queueing are separate cost categories. Hardware and native input sizes can differ across adapters, so cross-model seconds do not by themselves establish algorithmic speed rankings. For the shared hinge-step300 target, the first observed reaching points are Qwen hinge/BAL 300/250 and InternVL 300/300. The accompanying CSV retains all recorded thresholds, including unreached targets; the figure connects evaluated points only.
% 
\begin{table}[t]
\centering\small
\caption{Measured 250-update WASP generation time (seconds per image), averaged over newly computed successful attacks only. Reused attacks are excluded.}
\label{tab:runtime250}
\begin{tabular}{llrr}
\toprule
Dataset & Model & Images & Mean (s) \\
\midrule
DocVQA & qwen & 1163 & 1040.9\\
DocVQA & internvl & 1086 & 469.9\\
TextVQA & qwen & 882 & 215.6\\
TextVQA & internvl & 823 & 311.3\\
\bottomrule
\end{tabular}
\end{table}

The runtime means use successful, newly computed BAL attacks only: 1,163/1,086 DocVQA images and 882/823 TextVQA images for Qwen/InternVL. Historical attacks reused by identity are excluded, rather than assigned zero cost or rescaled from 300 steps. Each timer includes image loading and native preparation, clean-feature encoding, objective setup, all 250 updates, and final PNG/backup export. It excludes model loading, precomputed layout generation, downstream QA, judging, and queueing. These are mean per-image generation times, not dataset wall-clock times; the model-specific subsets and native image sizes differ.

\begin{table}[t]
\centering\small
\caption{Auxiliary 200-page study: loss-selected image statistics from 300-update attacks. Area is actual stroke-union percentage; PSNR is relative to clean (dB); RMS is within-mask RGB change relative to plain on $[0,1]$.}
\label{tab:visual}
\setlength{\tabcolsep}{4pt}
\begin{tabular}{lrrr}
\toprule
Model & Area & PSNR & RMS \\
\midrule
Qwen2.5-VL & 7.995 & 23.43 & 0.1209\\
InternVL2.5 & 7.995 & 21.90 & 0.2017\\
\bottomrule
\end{tabular}
\end{table}

\paragraph{Distortion definitions.}
For final PNG $x_a$ and plain PNG $x_w$, within-mask RMS is
\begin{equation}
 \sqrt{\frac{\sum_{p,c} M_p(x_{a,pc}-x_{w,pc})^2}{3|M|}}.
\end{equation}
This statistic divides by three RGB channels, unlike the training regularizer's squared-error term. Whole-page PSNR is measured relative to clean pixels and is averaged per page. The original 200-page visual reports contain zero mask-outside change counts. The earlier Qwen/InternVL audit rebuilt 183 of 2,400 final images from saved parameters. The reconstruction audit covers this 183-image subset.

\section{Parsing and Metric Coverage}
The parsing experiment targets the layout-stage image representations and evaluates the complete MinerU2.5 pipeline. Discrete layout generation, routing, crop choice, and document assembly are evaluated through the real pipeline rather than treated as exactly differentiable.

For the full clean/plain/attack evaluation, official per-page ED aggregation uses 1,557 text pages, 313 formula pages, and 1,638 reading-order pages. TEDS uses 665 table instances and CDM uses 2,352 formula instances. All 1,651 input pages are retained under the official category-missing rules. A verified attack-induced native parser failure is evaluated as empty output; infrastructure failures are not attack successes. The parser attack optimizes layout-stage visual representations without independently optimizing recognition crops.

For the routing diagnostic, 205 OmniDocBench pages were sampled uniformly without replacement (seed 20261028); the first five were reserved for development and the remaining 200 for evaluation. On this random 200-page subset, the 300-update route-swapping diagnostic compares the complete attacked run, attacked image content recognized using clean routing, and clean content recognized using attacked routing. The corresponding text ED values are 0.846946, 0.071535, and 0.842905. The clean and plain full-pipeline values are 0.062433 and 0.055212. Together with 37 empty attacked layouts and reduced content coverage, the routing swaps identify upstream layout and routing as a major failure pathway in the evaluated parser. The conditions compare complete pipeline outcomes.

\section{Carrier Geometry and Text Overlap}
\begin{table*}[t]
\centering\small
\caption{Full carrier measurements on InternVL. All configurations have 3146.24 visual cells per page on average, including local tiles and thumbnails. Rates are means of per-page fractions, not ratios of dataset-level sums. Area, Touch, Cov$_{10}$ and ASR are percentages; counts are page means.}
\label{tab:fullcoverage}
\setlength{\tabcolsep}{4pt}
\begin{tabular}{lrrrrrrrrr}
\toprule
Word & Font & Area & Touch & Cov$_{10}$ & $N_{touch}$ & $N_{10}$ & Plain & WASP & ANLS \\
\midrule
confidential & 24 & 1.995 & 18.26 & 6.29 & 572.4 & 197.2 & 2.70 & 23.46 & 0.7225\\
confidential & 24 & 3.994 & 34.03 & 13.46 & 1066.9 & 423.2 & 5.65 & 44.23 & 0.5448\\
confidential & 24 & 7.995 & 61.62 & 27.31 & 1932.1 & 856.2 & 9.46 & 71.99 & 0.2933\\
confidential & 24 & 9.996 & 70.41 & 39.49 & 2209.4 & 1240.3 & 13.76 & 77.64 & 0.2220\\
confidential & 16 & 7.998 & 76.47 & 30.89 & 2404.6 & 973.6 & 7.74 & 66.83 & 0.3347\\
confidential & 36 & 7.992 & 48.38 & 25.11 & 1514.3 & 788.3 & 9.58 & 71.01 & 0.2892\\
copyright & 24 & 7.996 & 65.58 & 28.43 & 2055.5 & 893.4 & 10.44 & 72.85 & 0.2856\\
\bottomrule
\end{tabular}
\end{table*}

Let $\widetilde M^{(v)}$ denote the binary glyph mask mapped to native view $v$ by bicubic interpolation and clipped to $[0,1]$. For each connector-associated $28\times28$ input cell $C_i$, we compute
\begin{equation}
 r_i=\frac{1}{|C_i|}\sum_{p\in C_i}\widetilde M^{(v)}(p),\quad
 P_{touch}=\frac{1}{N}\sum_i\mathbf{1}[r_i>0].
\end{equation}
Similarly, $P_{10}=N^{-1}\sum_i\mathbf{1}[r_i\geq0.10]$. The mapped mask is not weighted by watermark opacity. Both local and thumbnail cells contribute to $N$. Each rate is calculated per page before averaging; a small nonzero resampled value counts as touched, making the 10\% threshold a useful companion statistic. Coverage measures the geometric reach of editable strokes in the input grid. It does not measure attention, feature corruption, or an isolated receptive field after token interaction. Plain and optimized images have identical coverage because optimization never changes $M$.

Across 2\%, 4\%, 8\%, and 10\% area at 24 pixels, touched-cell rates are 18.26\%, 34.03\%, 61.62\%, and 70.41\%; coverage at the 10\% threshold is 6.29\%, 13.46\%, 27.31\%, and 39.49\%. Increasing area increases both geometric reach and ASR in these settings. The fixed-area font comparison, however, provides a counterexample to interpreting this association as a general rule: the most widely touching font has the weakest attack among the three tested sizes. Pairwise intervals use the same page-level bootstrap protocol as the task comparisons.

At 24 pixels and 8\% area, replacing confidential with copyright gives 72.85\% ASR versus 71.99\%, a difference of 0.86 points with CI $[-2.54,4.36]$. Plain ASR is 10.44\% versus 9.46\%, while cell-touch coverage rises to 65.58\%. The two tested words both support strong attacks, with comparable ASR. Changing wording also changes character count and stroke geometry.

For the 8\%-to-10\% area increment, ANLS decreases from 0.2933 to 0.2220, while Plain ASR increases from 9.46\% to 13.76\%. At 10\%, optimization adds 63.88 ASR points over Plain. At step 100, ASR is 14.25/32.56/57.62/65.36\% for 2/4/8/10\% area; these fixed-step measurements differ from the loss-selected endpoints.

\paragraph{Three distinct notions of coverage.}
Stroke area $|M|/(HW)$ measures the portion of the page that is editable. Visual-cell coverage measures how those strokes intersect the frontend's input grid. Source-text overlap instead asks how much existing page foreground lies under the strokes. For a fixed clean-page dark-foreground set $D$, the corresponding statistic is
\begin{equation}
 O_{text}=\frac{|M\cap D|}{|D|},\qquad |D|>0.
\end{equation}
Dark foreground is a text proxy: it can include rules, graphics, and scan artifacts, and is not equivalent to OCR text boxes or glyph-level text annotations. This geometric quantity is identical for plain and optimized versions sharing the same $M$ and clean-page $D$. Consequently, their task-performance gap cannot be explained by additional spatial coverage of source foreground. Optimization changes pixel values within that shared support. We assess its perceptual effect separately through the human readability study; $O_{text}$ quantifies source-foreground overlap alongside page area and visual-cell coverage.

\section{Text-Region and Background Placement Controls}
\paragraph{Common candidates and constraints.}
Text-prioritized (Text) and background-prioritized (BG) layouts use the same clean-page geometry and complete glyph without resampling. Candidate origins lie on a 12-source-pixel grid. Candidates must remain inside the page, with nonoverlapping editable stroke supports and center separation of at least half the glyph's shorter bounding-box side. The target number is $K=\lfloor0.08HW/a_g\rfloor$, where $a_g$ is the editable area of one complete glyph. Both layouts retain identical $K$ and actual stroke area on each page; every page reaches its target. Discrete glyph counts give mean area 7.969\%; 189/200 pages lie in $[7.92\%,8\%]$. The full 200-page sample is primary, with a geometry-defined area sensitivity subset retained in the source tables. Uniform-layout results are separate because their instance geometry is not matched to these new layouts.

\paragraph{Preference maps and greedy selection.}
Let $T$ be the union of frozen clean MinerU text-class blocks and $O$ the union of all content blocks, with $B=\neg O$. Text classes include text, titles, lists, reference text, headers, footers, page numbers, and text captions/footnotes; figures, tables, and equations are not treated as background. For preference $P\in\{T,B\}$, rank each legal glyph by the fraction of its editable strokes in $P$. Select the highest score; ties prefer larger distance from existing centers, then smaller candidate ID. The common procedure removes conflicting placements and uses at most two bounded repairs with up to 64 alternatives each. If preferred regions are exhausted, legal placements elsewhere remain available; layout names indicate preference, not guaranteed exclusive occupancy. Table~\ref{tab:layout_geometry} verifies the resulting intervention.

\begin{table*}[t]
\centering\small
\caption{Geometry averaged equally over pages, in percentages. $P_T$/$P_B$ divide editable strokes in text/background regions by total stroke area; $O_D$/$O_T$ divide covered foreground by all foreground/text-restricted foreground. There are 200 pages except $O_T$, which has 198 eligible pages.}
\label{tab:layout_geometry}
\setlength{\tabcolsep}{4pt}
\begin{tabular}{lrrrrr}
\toprule
Placement & Area & $P_T$ & $P_B$ & $O_D$ & $O_T$ \\
\midrule
Text & 7.97 & 51.53 & 36.59 & 13.58 & 25.90 \\
BG & 7.97 & 1.36 & 95.44 & 2.99 & 1.32 \\
\bottomrule
\end{tabular}
\end{table*}

\paragraph{Foreground overlap and subsets.}
The source-foreground mask is $D=\{p:I_{gray}(p)<\operatorname{BoxBlur}_{15}(I_{gray})(p)-10\}$ in 8-bit intensity, with the frozen Pillow boundary rule. This experiment additionally measures foreground restricted to text blocks, $D_T=D\cap T$, and $O_T=|M\cap D_T|/|D_T|$. It differs from whole-page foreground overlap $|M\cap D|/|D|$ defined earlier. Empty $D_T$ yields an undefined value, not zero; there are 198 eligible pages for $O_T$. Low-overlap membership is frozen before downstream QA: the BG mask must place at least 80\% of strokes in $B$ and cover at most 5\% of $D_T$. This selects 182 pages and 839 questions, with 795 Qwen and 763 InternVL clean-correct answers. BG plain/random/WASP ASR is 1.26/1.34/98.99\% for Qwen and 3.28/2.97/19.53\% for InternVL. The strong Qwen attack persists on this low-overlap subset, supporting effectiveness beyond substantial direct text coverage. Overlap is measured using the stated block and foreground proxies.

The mean fraction of strokes within text blocks is 51.53\% for Text and 1.36\% for BG; BG places 95.44\% outside all detected content blocks. Mean text-block foreground coverage is 25.90\% versus 1.32\% on the 198 eligible pages. The median paired text-placement difference is 48.50 percentage points, with the intended ordering on 197/200 pages. Both layouts use 24-pixel confidential glyphs, $45^\circ$ rotation, opacity 0.30, and 300 updates with step 300 as the primary endpoint.

\paragraph{Random controls and aggregation.}
For each page, model, and layout, three fixed random streams (seeds 20261108--20261110) draw independent uniform $[-1,1]$ values per editable pixel, broadcast over RGB. The scalar amplitude is searched for at most 64 evaluations, within the same perturbation bound as WASP. Matching uses RMS relative to the corresponding plain PNG after clipping and 8-bit quantization, with 1\% relative tolerance, or absolute $1/255$ when the target RMS is below $1/255$. Outside-mask changes are zero. Of 2,400 controls, 2,397 match; all three exceptions belong to InternVL text preference. We retain them rather than silently filtering pages, so that arm's complete random mean is not a strictly matched-amplitude estimate. Both background arms and the Qwen text arm match for every control. Random-seed outcomes are averaged within page before paired inference. These controls are specific to their placement layouts and are separate from the measured uniform-layout Random controls in the main QA table.

\begin{table*}[t]
\centering\small
\caption{Complete paired-layout endpoints. EM and ASR are percentages. Both models use 894 questions; ASR denominators are 846/814 for Qwen/InternVL. Random averages three directions and retains the three unmatched InternVL Text controls. Best is selected by attack loss, not downstream answers.}
\label{tab:layout_endpoints}
\setlength{\tabcolsep}{4pt}
\begin{tabular}{lllrrr}
\toprule
Model & Placement & Input & ANLS & EM (\%) & ASR (\%) \\
\midrule
Qwen2.5-VL & Text & Plain & 0.8812 & 79.53 & 4.61 \\
Qwen2.5-VL & Text & Random & 0.8740 & 78.00 & 6.26 \\
Qwen2.5-VL & Text & WASP-300 & 0.0332 & 2.13 & 96.57 \\
Qwen2.5-VL & Text & WASP-Best & 0.0338 & 2.13 & 96.57 \\
Qwen2.5-VL & BG & Plain & 0.8986 & 83.33 & 1.42 \\
Qwen2.5-VL & BG & Random & 0.8983 & 83.04 & 1.65 \\
Qwen2.5-VL & BG & WASP-300 & 0.0045 & 0.45 & 99.05 \\
Qwen2.5-VL & BG & WASP-Best & 0.0045 & 0.45 & 99.05 \\
InternVL2.5 & Text & Plain & 0.8179 & 72.37 & 11.67 \\
InternVL2.5 & Text & Random & 0.8234 & 73.12 & 11.10 \\
InternVL2.5 & Text & WASP-300 & 0.3472 & 24.94 & 64.13 \\
InternVL2.5 & Text & WASP-Best & 0.3474 & 24.83 & 64.25 \\
InternVL2.5 & BG & Plain & 0.8784 & 81.21 & 4.05 \\
InternVL2.5 & BG & Random & 0.8802 & 81.62 & 3.48 \\
InternVL2.5 & BG & WASP-300 & 0.7079 & 60.63 & 21.62 \\
InternVL2.5 & BG & WASP-Best & 0.7116 & 60.74 & 21.38 \\
\bottomrule
\end{tabular}
\end{table*}

\begin{table*}[t]
\centering\small
\caption{Paired ASR contrasts at step 300 in percentage points; $\Delta$ denotes WASP minus the layout-specific Plain. Unadjusted 95\% percentile intervals use 10,000 page-cluster bootstrap resamples. BG random controls all meet the RMS-matching tolerance.}
\label{tab:layout_ci}
\setlength{\tabcolsep}{4pt}
\begin{tabular}{llrr}
\toprule
Model & Contrast & Difference & 95\% CI \\
\midrule
Qwen2.5-VL & Text: WASP $-$ Plain & 91.96 & [89.18, 94.48] \\
Qwen2.5-VL & BG: WASP $-$ Plain & 97.64 & [96.27, 98.77] \\
Qwen2.5-VL & BG: WASP $-$ Random & 97.40 & [96.19, 98.46] \\
Qwen2.5-VL & Text $\Delta$ $-$ BG $\Delta$ & -5.67 & [-8.09, -3.45] \\
InternVL2.5 & Text: WASP $-$ Plain & 52.46 & [47.79, 57.05] \\
InternVL2.5 & BG: WASP $-$ Plain & 17.57 & [13.85, 21.45] \\
InternVL2.5 & BG: WASP $-$ Random & 18.14 & [14.34, 22.12] \\
InternVL2.5 & Text $\Delta$ $-$ BG $\Delta$ & 34.89 & [29.06, 40.69] \\
\bottomrule
\end{tabular}
\end{table*}

\paragraph{Uncertainty and scope.}
Intervals use 10,000 paired page-cluster bootstrap resamples (seed 20261108), recomputing the model-specific clean-correct denominator in each resample. Table~\ref{tab:layout_ci} reports optimization increments and contrasts; positive increments mean greater attack damage. Intervals are unadjusted percentile intervals, without significance tests or multiplicity-adjusted claims. The position intervention changes spatial concentration and native input-cell coverage along with text overlap. It therefore tests the complete placement change, rather than isolating an individual token, attention process, or exact character-level causal effect.

\section{Cross-Model Transfer Protocol and Secondary Checks}
All transfers reuse unchanged source-optimized PNGs on 200 DocVQA pages / 894 questions, selecting the minimum source loss from 300-update runs without target optimization or target-based checkpoint selection. Target clean-correct denominators are 814/846/513/825/763 for InternVL-8B/Qwen-7B/Vicuna/Qwen-3B/InternVL-2B. Mistral and Vicuna denote the LLaVA-v1.6 7B variants. Confidence intervals use 10,000 page-paired bootstrap resamples.

InternVL-8B to InternVL-2B yields 50.07\% ASR versus 12.06\% for Plain (+38.01 points; 95\% CI $[33.83,42.20]$). Mistral to Vicuna yields 66.28\% versus 24.17\% (+42.11 points; $[37.23,46.95]$). Qwen-7B to Qwen-3B yields 17.09\% versus 6.06\% (+11.03 points; $[8.46,13.72]$). For the InternVL and Qwen scale pairs, target clean accuracies are 85.35\% and 92.28\%; gains remain 37.79 and 10.92 points on their source--target common clean-correct subsets (741 and 815 questions). The audited scale pairs have different visual weights and connector shapes despite matching vision-tower architectures. These exploratory comparisons reuse the auxiliary 200-page set and remain distinct from the full-validation, 250-update main experiments.

\section{Resizing and Scene-Text Extension}
\begin{table*}[t]
\centering\small
\caption{Resizing evaluation: 96 pages, 403 questions. Weighted training is an extension, not the default single-scale method.}
\label{tab:scale}
\setlength{\tabcolsep}{4pt}
\begin{tabular}{lllrr}
\toprule
Model & Input & Attack & ASR (\%) & ANLS \\
\midrule
Qwen2.5-VL & original & plain & 4.23 & 0.8577\\
Qwen2.5-VL & original & single:300 & 97.62 & 0.0211\\
Qwen2.5-VL & original & original\_weighted:300 & 85.98 & 0.1225\\
Qwen2.5-VL & cap1280 & plain & 9.09 & 0.8397\\
Qwen2.5-VL & cap1280 & single:300 & 16.58 & 0.7928\\
Qwen2.5-VL & cap1280 & original\_weighted:300 & 32.89 & 0.6679\\
Qwen2.5-VL & cap1600 & plain & 6.10 & 0.8644\\
Qwen2.5-VL & cap1600 & single:300 & 19.63 & 0.7583\\
Qwen2.5-VL & cap1600 & original\_weighted:300 & 22.81 & 0.7477\\
Qwen2.5-VL & cap1920 & plain & 7.65 & 0.8465\\
Qwen2.5-VL & cap1920 & single:300 & 25.33 & 0.6731\\
Qwen2.5-VL & cap1920 & original\_weighted:300 & 40.90 & 0.5800\\
InternVL2.5 & original & plain & 9.42 & 0.8146\\
InternVL2.5 & original & single:300 & 72.02 & 0.2925\\
InternVL2.5 & original & original\_weighted:300 & 69.25 & 0.2927\\
InternVL2.5 & cap1280 & plain & 10.83 & 0.8096\\
InternVL2.5 & cap1280 & single:300 & 48.89 & 0.5285\\
InternVL2.5 & cap1280 & original\_weighted:300 & 63.89 & 0.3641\\
InternVL2.5 & cap1600 & plain & 10.44 & 0.8067\\
InternVL2.5 & cap1600 & single:300 & 62.91 & 0.3974\\
InternVL2.5 & cap1600 & original\_weighted:300 & 66.48 & 0.3250\\
InternVL2.5 & cap1920 & plain & 10.50 & 0.8077\\
InternVL2.5 & cap1920 & single:300 & 67.40 & 0.3358\\
InternVL2.5 & cap1920 & original\_weighted:300 & 67.68 & 0.3126\\
\bottomrule
\end{tabular}
\end{table*}

The resizing study first screens strategies on 32 pages and evaluates the selected configuration on a 96-page, 403-question confirmation subset. The original single-scale method and weighted multiscale extension remain separate conditions. The additional Qwen weighted experiment is a post-hoc extension; it improves capped-scale average ASR but sacrifices original-resolution performance. Page counts, transformed-subset membership, and target-specific clean-correct denominators are retained in the source tables rather than pooled with the main DocVQA evaluation.

\begin{table}[t]
\centering\small
\caption{TextVQA, 250 updates, using the same 1,000-image / 1,592-question validation sample as the main table. Paired ASR differences in percentage points; 10,000 image-bootstrap resamples, with all questions and random seeds kept together.}
\label{tab:textvqa}
\begin{tabular}{llrr}
\toprule
Model & Contrast & $\Delta$ASR & 95\% CI \\
\midrule
Qwen & WASP--Plain & 52.63 & [49.69, 55.63]\\
Qwen & WASP--Random & 52.33 & [49.44, 55.29]\\
InternVL & WASP--Plain & 48.47 & [45.67, 51.27]\\
InternVL & WASP--Random & 47.50 & [44.71, 50.21]\\
\bottomrule
\end{tabular}
\end{table}

% BEGIN retained DocOwl table (not rendered)
% \begin{table*}[t]
% \centering\small
% \caption{TextVQA: 200 images, 316 questions. The task score is official VQA accuracy. The condition identifies the source model; both endpoints are retained. DocOwl2 cells are reserved for pending evaluation.}
% \label{tab:textvqa}
% \setlength{\tabcolsep}{4pt}
% \begin{tabular}{llrrr}
% \toprule
% Target & Condition & Success / eligible & ASR (\%) & VQA accuracy \\
% \midrule
% Internvl & T8:internvl:best\_loss & 153/268 & 57.09 & 0.1883\\
% Internvl & T8:internvl:step300 & 151/268 & 56.34 & 0.1915\\
% Internvl & T8:qwen:best\_loss & 40/268 & 14.93 & 0.6177\\
% Internvl & T8:qwen:step300 & 39/268 & 14.55 & 0.6177\\
% Internvl & clean & 0/268 & 0.00 & 0.7259\\
% Internvl & plain & 27/268 & 10.07 & 0.6585\\
% Qwen & T8:internvl:best\_loss & 38/274 & 13.87 & 0.6491\\
% Qwen & T8:internvl:step300 & 38/274 & 13.87 & 0.6513\\
% Qwen & T8:qwen:best\_loss & 178/274 & 64.96 & 0.2405\\
% Qwen & T8:qwen:step300 & 176/274 & 64.23 & 0.2449\\
% Qwen & clean & 0/274 & 0.00 & 0.7468\\
% Qwen & plain & 16/274 & 5.84 & 0.7222\\
% \midrule
% DocOwl2 & clean & & & \\
% DocOwl2 & plain & & & \\
% DocOwl2 & T8:docowl2:best\_loss & & & \\
% DocOwl2 & T8:docowl2:step300 & & & \\
% \bottomrule
% \end{tabular}
% \end{table*}
% END retained DocOwl table

TextVQA uses a fixed 1,000-image validation sample (1,592 questions), formed by retaining the verified random 200-image sample and sampling 800 additional unique images with seed 20261112. Both models perform 250 updates and select the minimum complete loss within updates 1--250. Both models use the same image and question identities across conditions. Random averages three seeds per question; bootstrap resampling keeps all questions and seeds of each image together. TextVQA is an open-ended short-answer task without answer choices. Its official VQA accuracy scores generated answers against annotated human responses; DocVQA is reported separately with ANLS and EM. Each source model is optimized separately on scene-text images. DocVQA and TextVQA form the two text-centric QA benchmarks in the main evaluation. Attacks are generated separately for each dataset and source model.
% Retained prior transfer statement:
% TextVQA uses 200 images and 316 questions, with clean-correct denominators 274 for Qwen and 268 for InternVL. TextVQA is an open-ended short-answer task without answer choices. Its official VQA accuracy scores generated answers against annotated human responses; DocVQA is reported separately with ANLS and EM. Each source model is optimized separately on scene-text images; cross-model transfer reuses the corresponding source PNG. DocVQA and TextVQA form the two text-centric QA benchmarks in the main evaluation. Attacks are generated separately for each dataset and source model.

\section{Possible Explanations for Model-Dependent Behavior}
\label{sec:model_behavior}
\paragraph{Observed patterns and scope.}
Two comparisons show different behavior on Qwen and InternVL. In the full-validation, 250-update aggregation ablation, BAL increases selected-loss ASR by 9.21 percentage points on Qwen and 2.24 points on InternVL, with paired 95\% intervals of $[7.68,10.76]$ and $[1.14,3.37]$, respectively. BAL therefore improves final ASR on both models, with a larger gain on Qwen. In the separate 200-page, 300-update placement experiment, Qwen reaches 96.57\% ASR with text preference and 99.05\% with background preference; InternVL reaches 64.13\% and 21.62\%. The text-minus-background difference in the WASP-minus-Plain increment is $-5.67$ points for Qwen and $34.89$ points for InternVL. These experiments use different budgets and supports. Their shared model dependence motivates the architectural hypotheses below, which require targeted interventions to test.

\paragraph{Regional averaging and BAL.}
Global token averaging weights each region in proportion to its token count. A small region can therefore retain high clean-feature alignment while contributing little to the global objective. BAL first averages within each group and then emphasizes groups with higher residual alignment. Qwen's clean-layout partitions may make this reweighting more closely correspond to localized document content, including small text regions whose surviving information could support an answer. InternVL's groups instead correspond to local tiles and the thumbnail: a tile can contain multiple text regions and background, so group-level reweighting still averages heterogeneous content within that tile. This difference may help explain BAL's larger gain on Qwen. Group weights follow residual feature alignment, allowing optimization without questions or answer-region annotations.

\paragraph{Placement and visual interactions.}
Qwen represents the page on a visual token grid with cross-region interactions, providing a possible route for background perturbations to alter representations of content elsewhere on the page. InternVL encodes local tiles and a thumbnail before language-model processing; perturbations near text may more directly affect the local views that carry readable evidence. Content can also occur in both a local tile and the thumbnail, potentially providing alternative visual evidence. Such redundancy offers one possible explanation for InternVL's lower sensitivity to background placement. Direct measurements of cross-region interactions would help test this account. Background preference retains some text overlap; the low-overlap subset nevertheless preserves the qualitative model contrast.

\paragraph{Empirical gains and mechanistic scope.}
The paired ablation establishes improved final ASR under BAL for both models at the common 250-update budget. The trajectory study additionally shows stronger intermediate attacks on Qwen, while InternVL has smaller differences and lower BAL ASR at some early checkpoints. These observations distinguish final effectiveness from the optimization path. The ablation compares each arm's complete optimization-and-selection procedure, with initial gradient-scale calibration matching median starting magnitudes. Architecture, native preprocessing, grouping, and optimization dynamics vary across models; controlled interventions on these components would test the proposed regional-averaging and multi-view explanations.

\section{Qualitative Selection and Provenance}
The original 300-update DocVQA example in the qualitative figure of the main paper is Qwen question 37370, page 10759, from the clean-correct, wrong-key stratum under the BAL selected-loss attack. Candidates with complete answers under 110 characters are sorted by question ID and shuffled with seed 20261107; the first inspected candidate is retained with frozen crop coordinates. A previously illustrated still-correct example (question 18762, page 5732) remains in the source archive but is not displayed in the revised figure.

The original 300-update OmniDocBench example is \texttt{PPT\_CalculusReview\_page\_014.png}, parsed by MinerU2.5. Candidates are PPT pages with clean text ED below 0.15 and attacked text ED above 0.8; we inspect for a recognizable formula and verify the empty full parsing output. This page has clean/attacked text ED of 0/1. The displayed clean formula is a typeset excerpt of the saved Markdown, whereas the attacked output is empty for the entire page. Complete clean/attacked images, Markdown outputs, ground truth, SHA256 identities, and crop coordinates accompany the figure sources. Both examples are selected illustrations rather than random samples or statistical evidence of human legibility.

\section{Human Readability Evaluation}
\label{sec:human_protocol}
\paragraph{Participants and scope.}
Fifteen volunteers completed 200 image ratings each, yielding 3,000 ratings across Clean, Plain, WASP--Qwen, and WASP--InternVL, with 750 ratings per condition. The study assesses perceived readability of source document and watermark content in the QA image conditions; QA correctness and exact transcription are separate outcomes.

\paragraph{Recruitment and privacy.}
Participants were volunteers recruited through faculty contacts within the university. We report readability ratings only in aggregate and do not disclose participants' names or other identifying details.

\paragraph{Assignment and presentation protocol.}
The study used 200 source pages with four image conditions. Each participant was assigned one version of each source page, with 50 images per condition, avoiding repeated exposure to different versions of the same page. For page index $i=0,\ldots,199$ and participant index $j=0,\ldots,14$, the balanced assignment was $c_{ij}=(i+j)\bmod4$. Each page--condition pair consequently has three or four ratings. Presentation was independently shuffled, model/attack labels were concealed, and no side-by-side clean reference or model answers were shown. Whole-page inspection with zooming and panning was permitted, with no OCR assistance or per-image time limit and breaks allowed between blocks of 50 images. The two questions answered for each image are separate outcomes from the same rating, not independent observations. The study followed this protocol without deviations. Detailed sampling and session records are not included in the aggregate data snapshot.

\paragraph{Questions and interpretation.}
The document question asks whether all original document text, numbers, and symbols can be clearly read after inspecting the entire page, allowing zooming. It concerns visual distinguishability of characters, not subject-matter comprehension, and excludes the overlaid watermark from the definition of original content. The response options are \emph{Yes, all are readable}, \emph{No, some are unreadable}, and \emph{Cannot judge}.

The watermark question asks whether the complete wording of at least one fully visible watermark instance can be read, ignoring copies cut off by the page boundary. Responses are \emph{Yes}, \emph{No}, \emph{Cannot judge}, and \emph{No watermark is visible}. The criterion concerns at least one complete instance, not every repeated copy. Clean has no applicable watermark-readability outcome. Neither question requires a transcription or a document QA answer.

\paragraph{Reported aggregation.}
The reported results use pooled ratings within each condition. For condition $c$, let $N_c=750$ and let $y^{\mathrm{doc}}_{cj}$ and $y^{\mathrm{mark}}_{cj}$ indicate an affirmative response on rating $j$. We report
\begin{align}
 \mathrm{Doc.}_c&=100\frac{\sum_{j=1}^{N_c}y^{\mathrm{doc}}_{cj}}{N_c},\\
 \mathrm{Mark}_c&=100\frac{\sum_{j=1}^{N_c}y^{\mathrm{mark}}_{cj}}{N_c}.
\end{align}
Only affirmative responses enter the numerator; other completed responses do not. Administrative missingness is distinct from a completed non-affirmative response. The supplied results contain 750 completed ratings per condition. Percentages are rounded to one decimal place, and the table retains exact numerator/denominator counts. For example, WASP--Qwen document readability is $734/750=97.866\ldots\%$, reported as 97.9\%. ``Ratings'' counts image evaluations; $n=15$ denotes distinct volunteers. Clean watermark readability is N/A.

This pooled estimator gives equal weight to ratings, so page--condition cells with four ratings receive more weight than cells with three. It replaces the page-equal averaging proposed in the initial design. The available aggregate counts support pooled affirmative percentages; page-equal estimates, majority-vote outcomes, and non-affirmative subcategories require more detailed records.

\paragraph{Results and limitations.}
Document affirmative counts are 744, 738, 734, and 732 out of 750 for Clean, Plain, WASP--Qwen, and WASP--InternVL, respectively. Watermark affirmative counts are 717, 711, and 707 for Plain and the two attacked conditions. Relative to Plain, the Qwen and InternVL document rates are lower by $100(738-734)/750=0.53$ and $100(738-732)/750=0.80$ percentage points. The watermark decreases are 0.80 and 1.33 points. These descriptive differences are calculated from unrounded counts.

The results show high perceived readability under both attacked conditions, with document rates within 0.80 percentage points and watermark rates within 1.33 points of Plain. These are descriptive comparisons of self-reported readability. The manuscript data snapshot contains aggregate counts, so we report the observed differences without equivalence claims, significance tests, or confidence intervals. Participant/page-linked records would enable uncertainty estimates accounting for repeated ratings and the originally planned page-equal sensitivity analysis.

\end{document}